\documentclass[10pt,journal,compsoc]{IEEEtran}
\usepackage{xcolor}
\usepackage{amsfonts, amsmath, amssymb, amsthm}
\newtheorem{theorem}{Theorem}
\newtheorem{lemma}[theorem]{Lemma}

\theoremstyle{definition}
\newtheorem{definition}{Definition}
\newtheorem{remark}{Remark}

\usepackage{graphicx}
\usepackage{subcaption}
\usepackage{float}
\usepackage{tabularx}

\usepackage[colorlinks=true,linkcolor=black,citecolor=black,urlcolor=blue]{hyperref}
\usepackage{algorithm}
\usepackage{algorithmic}
\usepackage{subcaption}

\DeclareMathOperator*{\argmin}{arg\,min}

\ifCLASSOPTIONcompsoc
  \usepackage[nocompress]{cite}
\else
  \usepackage{cite}
\fi
\ifCLASSINFOpdf
\else
\fi
\begin{document}
%
\title{Discovering Patterns from Time-Varying Spatiotemporal Data with Low-Rank Autoregression}

\title{Discovering Dynamic Patterns from Spatiotemporal Data with Time-Varying Low-Rank Autoregression}
%
%
%
%

\author{Xinyu~Chen\textsuperscript{\textsection},~\IEEEmembership{Student Member,~IEEE,}
    Chengyuan~Zhang\textsuperscript{\textsection},~\IEEEmembership{Student Member,~IEEE,}
        Xiaoxu~Chen,
        Nicolas~Saunier,
        and~Lijun~Sun,~\IEEEmembership{Senior Member,~IEEE}
\IEEEcompsocitemizethanks{\IEEEcompsocthanksitem Xinyu Chen and Nicolas Saunier are with the Civil, Geological and Mining Engineering Department, Polytechnique Montreal, Montreal, QC
H3T 1J4, Canada. E-mail: chenxy346@gmail.com (Xinyu Chen), nicolas.saunier@polymtl.ca (Nicolas Saunier).
\IEEEcompsocthanksitem Chengyuan Zhang, Xiaoxu Chen, and Lijun Sun are with the Department of Civil Engineering, McGill University, Montreal, QC H3A 0C3, Canada. E-mail: enzozcy@gmail.com (Chengyuan Zhang), xiaoxu.chen@mail.mcgill.ca (Xiaoxu Chen), lijun.sun@mcgill.ca (Lijun Sun).
\IEEEcompsocthanksitem \textsuperscript{\textsection}Xinyu Chen and Chengyuan Zhang contributed equally to this work.
\protect\\
}
}

\IEEEtitleabstractindextext{%
\begin{abstract}
The problem of broad practical interest in spatiotemporal data analysis, i.e., discovering interpretable dynamic patterns from spatiotemporal data, is studied in this paper. Towards this end, we develop a time-varying reduced-rank vector autoregression (VAR) model whose coefficient matrices are parameterized by low-rank tensor factorization. Benefiting from the tensor factorization structure, the proposed model can simultaneously achieve model compression and pattern discovery. In particular, the proposed model allows one to characterize nonstationarity and time-varying system behaviors underlying spatiotemporal data. To evaluate the proposed model, extensive experiments are conducted on various spatiotemporal data representing different nonlinear dynamical systems, including fluid dynamics, sea surface temperature, USA surface temperature, and NYC taxi trips. Experimental results demonstrate the effectiveness of modeling spatiotemporal data and characterizing spatial/temporal patterns with the proposed model. In the spatial context, the spatial patterns can be automatically extracted and intuitively characterized by the spatial modes. In the temporal context, the complex time-varying system behaviors can be revealed by the temporal modes in the proposed model. Thus, our model lays an insightful foundation for understanding complex spatiotemporal data in real-world dynamical systems. The adapted datasets and Python implementation are publicly available at \url{https://github.com/xinychen/vars}.
\end{abstract}

}

\maketitle

\IEEEdisplaynontitleabstractindextext

%
\IEEEpeerreviewmaketitle

\section{Introduction}

\IEEEPARstart{D}{ynamic} mechanisms that drive nonlinear systems are universally complex and obscure. Straightforwardly, one can investigate the behavior of a system by discovering the patterns from its observations. In practice, when we take observations from a real-world complex system, spatiotemporal data are one of the most widely encountered form relating to space and time and showing the characteristics of time series. With the remarkable development of sensing technologies, tremendous spatiotemporal data are accessible for analysis, such as satellite data \cite{reynolds2002improved}, station-level weather observation data \cite{Daymet}, and trajectory of human mobility \cite{zheng2015trajectory}, to name but a few. In the literature, various spatiotemporal data were characterized by time series models. Leveraging time series models not only allows one to analyze spatiotemporal data but also makes it possible to discover inherent spatial and temporal patterns from the data over space and time.

As a simple yet efficient and classical statistical model, vector autoregression (VAR) explicitly finds the linear relationship among a sequence of time series changing over time, which can also successfully describe the dynamic behavior of time series \cite{hamilton1994time, lutkepohl2005new}. Formally, if the given time series $\boldsymbol{s}_{1},\ldots,\boldsymbol{s}_{T}\in\mathbb{R}^{N}$ (i.e., a collection of observations with $N$ variables at consecutive times) are stationary, then the $d$th-order VAR takes a linear formulation as $\boldsymbol{s}_{t}=\sum_{k=1}^{d}\boldsymbol{A}_{k}\boldsymbol{s}_{t-k}+\boldsymbol{\epsilon}_{t},\forall t\in\{d+1,\ldots,T\}$, in which the coefficient matrices $\boldsymbol{A}_{1},\ldots,\boldsymbol{A}_{d}\in\mathbb{R}^{N\times N}$ are expected to capture the temporal correlations of the multivariate time series, and $\boldsymbol{\epsilon}_{t}$ is the residual at time $t$. A large body of VAR and its variants such as reduced-rank VAR \cite{izenman1975reduced,ahn1988nested,velu1998multivariate}, dynamic mode decomposition (DMD) \cite{schmid2010dynamic, tu2013dynamic, kutz2016dynamic}, high-dimensional VAR \cite{wang2021high}, and time-varying VAR with tensor factorization \cite{harris2021time} have been developed for analyzing real-world time series data. Essentially, the modeling intuition of these data-driven approaches is that the coefficient matrix in VAR is rank-deficient, or at least the dominant patterns underlying the coefficient matrix can be revealed by a low-rank structure.

DMD models---the classical variants of VAR---were initially developed for discovering spatiotemporal coherent patterns from fluid flow mechanism, and they are available for interpreting the system behavior and underlying data patterns, giving credit to its low-rank structure (i.e., a reduced-rank linear system of VAR) for dimensionality reduction \cite{schmid2010dynamic, tu2013dynamic,kutz2016dynamic}. However, stationary data are often required when analyzing multivariate time series via VAR and its variants, which is against the fact that real-world data are usually nonstationary. Therefore, one great challenge of modeling time series with VAR is identifying the time-varying system behaviors in the analysis, which is often associated with the nonstationarity issue. 
Although the nonstationarity and time-varying system behaviors are pretty clear to verify, the problem of discovering underlying data patterns from time-varying systems is challenging and still demands further exploration. To characterize the underlying time-varying system behaviors hidden behind time series, a classical framework was proposed by utilizing the dynamic linear models (e.g., time-varying vector autoregression) whose latent variables vary over time \cite{west1997bayesian, primiceri2005time, del2015time}.

Typically, time-varying VAR model takes a sequence of VAR processes at different times, and it is capable of handling the time-varying system behaviors. Nevertheless, the time-varying coefficients in the model give rise to the over-parameterization issue, which implies that the number of parameters (e.g., coefficient matrices in time-varying VAR) inevitably exceeds the number of observations. The over-parameterized coefficients can be regularized by a certain smoothing term from a machine learning perspective, but it does not work in the high-dimensional settings. Another efficient approach to address the over-parameterization issue is using low-rank models such as low-rank matrix/tensor factorization for pursuing perfect model compression. Preferably, low-rank models have shown unprecedented compression power on both matrix and tensor data or parameters in several contemporary studies \cite{donoho2006compressed, eldar2012compressed, kolda2009tensor, anandkumar2014tensor, wright2022high}.

Complementary to the above methods, recent work provides an online DMD \cite{zhang2019online} and a time-varying autoregression \cite{harris2021time} for modeling time-varying systems. The online DMD model allows one to compute DMD in real time and find the approximation of a system's dynamics for time-varying streaming data. However, as for DMD models, they are sensitive and vulnerable to the noises in the time series data, posing both methodological and practical challenges \cite{wu2021challenges}. Therefore, it is hard to use DMD to discover data patterns and system behaviors from time-varying and noisy time series. In \cite{harris2021time}, the time-varying autoregression model has a sequence of windowed linear systems consisting of VAR, highly motivated by the need for time-varying system identification. It allows one to capture pattern changes of a dynamical system through tensor factorization. Typically, tensor factorization in the time-varying autoregression is capable of compressing the over-parameterized coefficients and identifying the time-varying system behaviors.

In this work, we revisit the classical spatiotemporal data analysis problem and introduce a novel method to discover spatial and temporal patterns from time-varying systems in a data-driven fashion. The cornerstone of this work is the time-varying VAR on nonstationary time series. Formally, we can formulate the first-order VAR with time-varying coefficients as $\boldsymbol{s}_{t}=\boldsymbol{A}_{t}\boldsymbol{s}_{t-1}+\boldsymbol{\epsilon}_{t},\forall t$, in which the system matrices $\{\boldsymbol{A}_{t}\}$ are changing over time. In this situation, the system matrices involve $\mathcal{O}(N^2(T-1))$ parameters, which exceed $\mathcal{O}(NT)$ observations. To address the over-parameterization issue, we utilize a low-rank tensor factorization structure---Tucker decomposition \cite{kolda2009tensor, golub2013matrix}---to achieve the model compression. In the meanwhile, tensor factorization allows one to discover spatiotemporal patterns from spatiotemporal data.

The main contributions of this work are twofold:
\begin{itemize}
    \item We propose a fully time-varying reduced-rank VAR model which allows one to discover interpretable dynamic modes from spatiotemporal data. In our model, we draw strong connections between time-varying VAR and tensor factorization (i.e., Tucker decomposition) that address the over-parameterization issue and reveal spatial and temporal patterns in the latent spaces.
    \item We demonstrate the model capability of the time-varying reduced-rank VAR for discovering interpretable patterns from extensive spatiotemporal datasets, including fluid dynamics, sea surface temperature, USA surface temperature, and NYC taxi trips. The evaluation results show that the latent variables in the tensor factorization of our model can reveal both spatial and temporal patterns. Time-varying system behaviors underlying these spatiotemporal data can be clearly identified by our model.
\end{itemize}

Focusing on the time-varying autoregression with low-rank structures, our model differs from the existing model proposed by \cite{harris2021time} on the technical side: i) the existing model takes a sequence of windowed (first-order) VAR processes, while our model is parameterized with fully time-varying coefficients in the higher-order VAR (see Definition~\ref{time_varying_var_definition} and Remark~\ref{windowed_model} for more details); ii) the existing model utilizes the CANDECOMP/PARAFAC (CP) decomposition (i.e., a special case of Tucker decomposition \cite{kolda2009tensor}) to circumvent the over-parameterized coefficients in which CP decomposition does not show any interactions between spatial modes and temporal modes, in contrast, our model applies the Tucker decomposition which allows us to better characterize spatiotemporal patterns. Without loss of generality, the proposed model lays a foundation for modeling real-world spatiotemporal data with the time-varying system behaviors and provides insights into dynamical system modeling.

The remainder of this work is organized as follows. Section~\ref{methodology} introduces a time-varying reduced-rank VAR model and presents an alternating minimization algorithm for solving the involved optimization problem. Section~\ref{experiment} shows extensive experiments on some real-world spatiotemporal datasets. We discuss this research and give a summary in Section~\ref{conclusion}. For more supplementary material, Appendix~\ref{appendix_dmd} presents spatial/temporal modes achieved by DMD on the fluid dynamics.

\section{Related Work}\label{related_work}



\subsection{Low-Rank Autoregression}

In the past decades, substantial studies on time series modeling have sought solutions to the over-parameterized autoregression, mainly aiming at model compression on the coefficients. Essentially, the idea of low-rank regression is taking advantage of dimensionality reduction techniques. To solve the over-parameterization issue in the high-dimensional VAR, there are some low-rank matrix and tensor tools ranging from the linear structure to multilinear structure, e.g., matrix factorization \cite{koop2019bayesian}, Tucker decomposition \cite{wang2021high}, and tensor train decomposition \cite{oseledets2011tensor}, to compress the coefficient matrices in VAR. At an early stage, a general reduced-rank setup for VAR models was proposed by Ahn and Reinsel \cite{ahn1988nested, reinsel1992vector}. Very recently, some representative models such as the multivariate autoregressive index model \cite{carriero2016structural} and Bayesian compressed VAR model \cite{koop2019bayesian} integrated matrix factorization into the framework. Another idea is to use a low-rank tensor structure to characterize the higher-order VAR (i.e., $d>1$), e.g., Tucker decomposition formatted VAR model in \cite{wang2021high}. Bahadori \emph{et al.} proposed a unified low-rank tensor learning framework for multivariate spatiotemporal analysis \cite{bahadori2014fast}. In these high-dimensional VAR models with the higher order, tensor factorization can perform better than matrix factorization in terms of model compression on the coefficients. However, tensor factorization would inevitably involve high complexity and nonconvex optimization. 
In the field of dynamical systems, DMD and higher-order DMD models can also be categorized as reduced-rank VAR models. They take the form of VAR and can handle high-dimensional time series because truncated singular value decomposition (SVD) and eigenvalue decomposition are adopted for preserving the most important dynamic modes of time series data \cite{schmid2010dynamic, tu2013dynamic, kutz2016dynamic, brunton2019data}. Therefore, matrix/tensor factorization in the low-rank autoregression not only addresses the over-parameterization issue, but also provides insights into pattern discovery.

\subsection{Time-Varying Autoregression}

In real-world applications, time series data are usually nonstationary. Recall that the stationary time series possess mean, average, and autocorrelation that are independent of time \cite{hamilton1994time}, but nonstationary time series violate that principle. Therefore, the nonstationarity characteristic poses great methodological and practical challenges for developing well-suited frameworks on such kind of data. Instead of stationarizing time series, the existing studies presented some autoregression models for characterizing time-varying system behaviors. In literature, a classical dynamic linear framework (e.g., time-varying VAR) utilizes time-varying latent variables to characterize the time-varying system behaviors of time series \cite{west1997bayesian,primiceri2005time,del2015time}. To reinforce such framework, many counterparts of time-varying autoregression such as time-varying structural VAR \cite{primiceri2005time,del2015time}, online DMD \cite{zhang2019online}, and time-varying autoregression with low-rank tensors \cite{harris2021time} have been introduced to time series analysis. Among these works, Harris \emph{et al.} \cite{harris2021time} introduced a time-varying VAR and addressed the over-parameterization issue through low-rank tensor factorization---tensor factorization allows one to achieve both model compression and pattern discovery. However, to the best of our knowledge, existing studies have not provided any systematical analysis for showing how to discover dynamic patterns from spatiotemporal data via time-varying autoregression.

\section{Methodology}\label{methodology}

This section introduces a fully time-varying reduced-rank VAR model for the multivariate time series. In particular, the model utilizes low-rank tensor factorization to compress the over-parameterized coefficients. Meanwhile, tensor factorization in the model allows us to automatically discover interpretable dynamic modes (corresponding to spatiotemporal patterns) from spatiotemporal data.


\subsection{Model Description}

In this work, we propose a time-varying VAR model with interpretability regarding the underlying system's dynamic behaviors in a data-driven fashion. This model provides a flexible framework built on the past efforts such as DMD \cite{schmid2010dynamic, kutz2016dynamic} and time-varying autoregression with low-rank tensors \cite{harris2021time}. For any observed spatiotemporal data in the form of multivariate time series, our model considers a time-varying linear system as follows.

\begin{definition}[Fully time-varying VAR]\label{time_varying_var_definition}
Given any multivariate time series data $\boldsymbol{S}\in\mathbb{R}^{N\times T}$ with columns $\boldsymbol{s}_{1},\ldots,\boldsymbol{s}_{T}\in\mathbb{R}^{N}$, if a $d$th-order VAR takes time-varying coefficients at time $t\in\{d+1,\ldots,T\}$, then the optimization problem of the fully time-varying VAR can be defined as follows,
\begin{equation}\label{opt_tv_var}
\min_{\{\boldsymbol{A}_{t}\}}~\frac{1}{2}\sum_{t=d+1}^{T}\|\boldsymbol{y}_{t}-\boldsymbol{A}_{t}\boldsymbol{z}_{t}\|_{2}^{2},
\end{equation}
where $\boldsymbol{z}_{t}\triangleq(\boldsymbol{s}_{t-1}^\top,\cdots,\boldsymbol{s}_{t-d}^\top)^\top\in\mathbb{R}^{dN}$ is an augmented vector and $\boldsymbol{y}_{t}\triangleq\boldsymbol{s}_{t}$. 
The data pair $\{\boldsymbol{y}_{t},\boldsymbol{z}_{t}\}$ are the inputs for learning the coefficient matrices $\{\boldsymbol{A}_{t}\in\mathbb{R}^{N\times (dN)}\}$. The notation $\|\cdot\|_{2}$ denotes the $\ell_{2}$-norm of vector. Notably, these coefficient matrices are capable of expressing the time series with a sequence of parameters.
\end{definition}

\begin{remark}\label{windowed_model}
In contrast to the fully time-varying VAR as mentioned in Definition~\ref{time_varying_var_definition}, Harris \emph{et al.} \cite{harris2021time} presented a windowed time-varying VAR with the first-order form (i.e., $d=1$), which is given by
\begin{equation}
\min_{\{\boldsymbol{A}_{t}\}}~\frac{1}{2}\sum_{t=1}^{M}\|\boldsymbol{Y}_{t}-\boldsymbol{A}_{t}\boldsymbol{Z}_{t}\|_{F}^2,
\end{equation}
where $\{\boldsymbol{A}_{t}\in\mathbb{R}^{N\times N}\}$ are the coefficient matrices. The time steps $T=MK,M\in\mathbb{N}^+,K\in\mathbb{N}^{+}$ is comprised of $M$ windows and each window is of length $K$. At any window $t$, the data pair $\{\boldsymbol{Y}_{t},\boldsymbol{Z}_{t}\}$ are defined as augmented matrices:
\begin{equation*}
\begin{aligned}
\boldsymbol{Y}_{t}&=\begin{bmatrix}
\mid & & \mid \\
\boldsymbol{s}_{(t-1)K+1} & \cdots & \boldsymbol{s}_{tK-1} \\
\mid & & \mid \\
\end{bmatrix}\in\mathbb{R}^{N\times (K-1)}, \\
\boldsymbol{Z}_{t}&=\begin{bmatrix}
\mid & & \mid \\
\boldsymbol{s}_{(t-1)K+2} & \cdots & \boldsymbol{s}_{tK} \\
\mid & & \mid \\
\end{bmatrix}\in\mathbb{R}^{N\times (K-1)}. \\
\end{aligned}
\end{equation*}
If $K=1$, then the windowed time-varying VAR is reduced to our fully time-varying VAR with $d=1$, showing the flexibility of the fully time-varying VAR over the windowed time-varying VAR.
\end{remark}

In Definition~\ref{time_varying_var_definition}, gathering $\{\boldsymbol{A}_{t}\}$ and stacking these matrices along an additional dimension, there exists a third-order tensor $\boldsymbol{\mathcal{A}}\in\mathbb{R}^{N\times (dN)\times(T-d)}$ for representing these coefficient matrices, in which $\boldsymbol{A}_{t}$ is the $t$th frontal slice of $\boldsymbol{\mathcal{A}}$. One great concern of the model is that it possesses $\mathcal{O}(dN^2(T-d))$ parameters, which would vastly exceed $\mathcal{O}(NT)$ observations in most cases, regardless of the order $d$ in the model. Therefore, one can demonstrate that, due to time-varying coefficients, the autoregression model involves the \emph{over-parameterization} issue. To address this issue, low-rank matrix/tensor factorization usually plays as fundamental tools for compressing the over-parameterized coefficients, e.g., matrix factorization in reduced-rank regression/autoregression \cite{izenman1975reduced, ahn1988nested, reinsel1992vector, velu1998multivariate} and tensor factorization in high-dimensional VAR \cite{wang2021high}. In this work, we assume that the tensor $\boldsymbol{\mathcal{A}}$ has a reduced rank, i.e., the rank of $\boldsymbol{\mathcal{A}}$ is deficient, and claim the existence of a multilinear rank-$(R,R,R)$ Tucker decomposition such that
\begin{equation}\label{tucker}
\boldsymbol{\mathcal{A}}=\boldsymbol{\mathcal{G}}\times_{1}\boldsymbol{W}\times_{2}\boldsymbol{V}\times_{3}\boldsymbol{X},
\end{equation}
where $\boldsymbol{\mathcal{G}}\in\mathbb{R}^{R\times R\times R}$ is the core tensor, and $\boldsymbol{W}\in\mathbb{R}^{N\times R},\boldsymbol{V}\in\mathbb{R}^{(dN)\times R},\boldsymbol{X}\in\mathbb{R}^{(T-d)\times R}$ are factor matrices. The notation $\times_{k}$ denotes the modal product between a tensor and a matrix along the mode $k$ \cite{kolda2009tensor}. If the autoregressive process is not time-varying, i.e., in the case of static coefficients, then the model would be reduced to the precedent reduced-rank VAR with matrix factorization \cite{ahn1988nested, reinsel1992vector, velu1998multivariate}. According to the property of Tucker decomposition and Kronecker product \cite{kolda2009tensor, golub2013matrix}, we can build connections between Eqs.~\eqref{opt_tv_var} and \eqref{tucker} through
\begin{equation}
\boldsymbol{A}_{t}=\boldsymbol{\mathcal{G}}\times_1\boldsymbol{W}\times_2\boldsymbol{V}\times_3\boldsymbol{x}_{t}^\top\equiv\boldsymbol{W}\boldsymbol{G}(\boldsymbol{x}_{t}^\top\otimes\boldsymbol{V})^\top,
\end{equation}
where $\boldsymbol{x}_{t}\in\mathbb{R}^{R}$ is the $t$th row of $\boldsymbol{X}$, and $\boldsymbol{G}\triangleq\boldsymbol{\mathcal{G}}_{(1)}\in\mathbb{R}^{R\times R^2}$ is the mode-1 unfolding/matricization of the core tensor $\boldsymbol{\mathcal{G}}$. The symbol $\otimes$ denotes the Kronecker product.

In this case, tensor factorization is used to compress the coefficients in the time-varying VAR. If one integrates the above-mentioned tensor factorization into time-varying VAR processes, then the optimization problem of the resultant time-varying reduced-rank VAR can be formulated as follows,
\begin{equation}\label{time_varying_model}
\min_{\boldsymbol{W},\,\boldsymbol{G},\,\boldsymbol{V},\,\boldsymbol{X}}~\frac{1}{2}\sum_{t=d+1}^{T}\|\boldsymbol{y}_{t}-\boldsymbol{W}\boldsymbol{G}(\boldsymbol{x}_{t}^\top\otimes\boldsymbol{V})^\top\boldsymbol{z}_{t}\|_{2}^{2}.
\end{equation}

As can be seen, we draw connections between Tucker decomposition and time-varying VAR and provide an expression for decomposing the coefficients into a sequence of matrices. The model not only addresses the over-parameterization issue in the time-varying VAR, but also provides physically interpretable modes for real-world time series data. As demonstrated by \cite{harris2021time}, one significant benefit of tensor factorization in the model is revealing the underlying data patterns in the parameters. For instance, on spatiotemporal data, $\boldsymbol{W}$ and $\boldsymbol{V}$ can be interpreted as spatial modes, while $\boldsymbol{X}$ can be interpreted as temporal modes. Time-varying $\{\boldsymbol{x}_{t}\}$ can explore the system's behavior of time series data. Since our model takes into account time-varying coefficients, we can offer a better understanding of dynamics of spatiotemporal data with nonstationarity and time-varying system behaviors. 
In contrast to our model, the time-varying autoregression with low-rank tensors builds VAR for the time series in each window and follows CP tensor factorization for model compression and outlier detection \cite{harris2021time}, and which seems to be less parsimonious than our model in terms of interpretable structures.

\begin{lemma}\label{matrix_form_opt_prop}

The optimization problem in Eq.~\eqref{time_varying_model} is equivalent to the following optimization problem in the form of matrix:
\begin{equation}  \label{eq: formulation}
\min_{\boldsymbol{W},\,\boldsymbol{G},\,\boldsymbol{V},\,\boldsymbol{X}}~\frac{1}{2}\|\boldsymbol{Y}-\boldsymbol{W}\boldsymbol{G}(\boldsymbol{X}\otimes\boldsymbol{V})^\top\tilde{\boldsymbol{Z}}\|_{F}^{2},
\end{equation}
where $\|\cdot\|_{F}$ denotes the Frobenius norm of matrix, and $\boldsymbol{Y},\tilde{\boldsymbol{Z}}$ are defined as follows,
\begin{equation*}
\boldsymbol{Y}\triangleq\begin{bmatrix}
\mid & \mid & & \mid \\
\boldsymbol{y}_{d+1} & \boldsymbol{y}_{d+2} & \cdots & \boldsymbol{y}_{T} \\
\mid & \mid & & \mid \\
\end{bmatrix}\in\mathbb{R}^{N\times(T-d)},
\end{equation*}
and
\begin{equation*}
\tilde{\boldsymbol{Z}}\triangleq\begin{bmatrix}
\boldsymbol{z}_{d+1} & \boldsymbol{0} & \cdots & \boldsymbol{0} \\
\boldsymbol{0} & \boldsymbol{z}_{d+2} & \cdots & \boldsymbol{0} \\
\vdots & \vdots & \ddots & \vdots \\
\boldsymbol{0} & \boldsymbol{0} & \cdots & \boldsymbol{z}_{T} \\
\end{bmatrix}\in\mathbb{R}^{(dN(T-d))\times (T-d)},
\end{equation*}
respectively.
\end{lemma}

\begin{remark}
This matrix-form optimization problem provides insights into understanding the problem of the time-varying reduced-rank VAR. Putting the orthogonal factor matrices with the matrix-form optimization problem together, our model can be written as a higher-order singular value decomposition (HOSVD, \cite{kolda2009tensor, golub2013matrix}) on the coefficient tensor. In this case, we give an unfolded form of HOSVD as follows,
\begin{equation}
\begin{aligned}
\min_{\boldsymbol{W},\,\boldsymbol{G},\,\boldsymbol{V},\,\boldsymbol{X}}~&\frac{1}{2}\|\boldsymbol{\mathcal{A}}_{(1)}-\boldsymbol{W}\boldsymbol{G}(\boldsymbol{X}\otimes\boldsymbol{V})^\top\|_{F}^{2} \\
\text{s.t.}~&\begin{cases}\boldsymbol{W}^\top\boldsymbol{W}=\boldsymbol{I}_{R},\\
\boldsymbol{V}^\top\boldsymbol{V}=\boldsymbol{I}_{R},\\
\boldsymbol{X}^\top\boldsymbol{X}=\boldsymbol{I}_{R},
\end{cases}
\end{aligned}
\end{equation}
where $\boldsymbol{\mathcal{A}}_{(1)}\triangleq\boldsymbol{Y}\tilde{\boldsymbol{Z}}^\dagger\in\mathbb{R}^{N\times (dN(T-d))}$ is the mode-1 unfolding of $\boldsymbol{\mathcal{A}}$. Herein, $\cdot^\dagger$ denotes the Moore-Penrose pseudo-inverse. In the constraint, $\boldsymbol{I}_{R}$ is the $R$-by-$R$ identity matrix.
\end{remark}

As can be seen, our model shows the capability of HOSVD for pattern discovery, but in high-dimensional settings, computing with the large and sparse matrix $\tilde{\boldsymbol{Z}}$ is computationally costly. In what follows, we let $f$ be the objective function of the optimization problem in Eq.~\eqref{time_varying_model}, and still use the vector-form formula to develop an alternating minimization algorithm for the time-varying reduced-rank VAR.

\subsection{Model Inference}

Recall that alternating minimization (e.g., alternating least squares) is a simple yet efficient method for addressing a manifold of nonconvex optimization problems in low-rank matrix/tensor factorization \cite{kolda2009tensor,  golub2013matrix}. To solve the involved optimization problem in our model, we apply an alternating minimization algorithm in which we have a sequence of least squares subproblems with respect to $\{\boldsymbol{W},\,\boldsymbol{G},\,\boldsymbol{V},\,\boldsymbol{X}\}$, respectively. Regarding the subproblem with respect to each unknown variable, we fix the remaining variables as known variables and intend to write down the least squares solutions or obtain the numerically approximated solutions. In addition, as reported in \cite{harris2021time}, the estimation based on alternating minimization has shown to be effective in such optimization problems.

\subsubsection{Updating the Variable $\boldsymbol{W}$}

With respect to the variable $\boldsymbol{W}$, the current estimation task is deriving closed-form solution from
\begin{equation}
\boldsymbol{W}:=\argmin_{\boldsymbol{W}}~\frac{1}{2}\sum_{t=d+1}^{T}\|\boldsymbol{y}_{t}-\boldsymbol{W}\boldsymbol{G}(\boldsymbol{x}_{t}^\top\otimes\boldsymbol{V})^\top\boldsymbol{z}_{t}\|_{2}^{2},
\end{equation}
while $\{\boldsymbol{G},\boldsymbol{V},\boldsymbol{X}\}$ are fixed as known variables. Since this optimization problem is convex, we can first write down the partial derivative of $f$ with respect to $\boldsymbol{W}$ as follows,
\begin{equation} \label{eq: derivate of W}
\begin{aligned}
\frac{\partial f}{\partial\boldsymbol{W}}=&-\sum_{t=d+1}^{T}(\boldsymbol{y}_{t}-\boldsymbol{W}\boldsymbol{G}(\boldsymbol{x}_{t}^\top\otimes\boldsymbol{V})^\top\boldsymbol{z}_{t})\boldsymbol{z}_{t}^\top(\boldsymbol{x}_{t}^\top\otimes\boldsymbol{V})\boldsymbol{G}^\top \\
=&-\sum_{t=d+1}^{T}\boldsymbol{y}_{t}\boldsymbol{z}_{t}^\top(\boldsymbol{x}_{t}^\top\otimes\boldsymbol{V})\boldsymbol{G}^\top \\
&+\sum_{t=d+1}^{T}\boldsymbol{W}\boldsymbol{G}(\boldsymbol{x}_{t}^\top\otimes\boldsymbol{V})^\top\boldsymbol{z}_{t}\boldsymbol{z}_{t}^\top(\boldsymbol{x}_{t}^\top\otimes\boldsymbol{V})\boldsymbol{G}^\top. \\
\end{aligned}
\end{equation}
Then let $\frac{\partial f}{\partial\boldsymbol{W}}=\boldsymbol{0}$, there exists a least squares solution to the variable $\boldsymbol{W}$:
\begin{equation}\label{least_square_w}
\begin{aligned}
\boldsymbol{W}=&\left(\sum_{t=d+1}^{T}\boldsymbol{y}_{t}\boldsymbol{z}_{t}^\top(\boldsymbol{x}_{t}^\top\otimes\boldsymbol{V})\boldsymbol{G}^\top\right) \\
&\cdot\left(\sum_{t=d+1}^{T}\boldsymbol{G}(\boldsymbol{x}_{t}^\top\otimes\boldsymbol{V})^\top\boldsymbol{z}_{t}\boldsymbol{z}_{t}^\top(\boldsymbol{x}_{t}^\top\otimes\boldsymbol{V})\boldsymbol{G}^\top\right)^{-1}.
\end{aligned}
\end{equation}

\subsubsection{Updating the Variable $\boldsymbol{G}$}

With respect to the variable $\boldsymbol{G}$, the estimation goal is finding the optimal solution from
\begin{equation}
\boldsymbol{G}:=\argmin_{\boldsymbol{G}}~\frac{1}{2}\sum_{t=d+1}^{T}\|\boldsymbol{y}_{t}-\boldsymbol{W}\boldsymbol{G}(\boldsymbol{x}_{t}^\top\otimes\boldsymbol{V})^\top\boldsymbol{z}_{t}\|_{2}^{2},
\end{equation}
while $\{\boldsymbol{W},\boldsymbol{V},\boldsymbol{X}\}$ are fixed as known variables. To find the closed-form solution from this optimization problem, the partial derivative of $f$ with respect to $\boldsymbol{G}$ is given by
\begin{equation}
\begin{aligned}
\frac{\partial f}{\partial\boldsymbol{G}}=&-\sum_{t=d+1}^{T}\boldsymbol{W}^\top(\boldsymbol{y}_{t}-\boldsymbol{W}\boldsymbol{G}(\boldsymbol{x}_{t}^\top\otimes\boldsymbol{V})^\top\boldsymbol{z}_{t})\boldsymbol{z}_{t}^\top(\boldsymbol{x}_{t}^\top\otimes\boldsymbol{V}) \\
=&-\sum_{t=d+1}^{T}\boldsymbol{W}^\top\boldsymbol{y}_{t}\boldsymbol{z}_{t}^\top(\boldsymbol{x}_{t}^\top\otimes\boldsymbol{V}) \\
&+\sum_{t=d+1}^{T}\boldsymbol{W}^\top\boldsymbol{W}\boldsymbol{G}(\boldsymbol{x}_{t}^\top\otimes\boldsymbol{V})^\top\boldsymbol{z}_{t}\boldsymbol{z}_{t}^\top(\boldsymbol{x}_{t}^\top\otimes\boldsymbol{V}). \\
\end{aligned}
\end{equation}
Let $\frac{\partial f}{\partial\boldsymbol{G}}=\boldsymbol{0}$, then there exists a least squares solution to the variable $\boldsymbol{G}$:
\begin{equation}\label{least_square_G}
\begin{aligned}
\boldsymbol{G}=&\boldsymbol{W}^\dagger\left(\sum_{t=d+1}^{T}\boldsymbol{y}_{t}\boldsymbol{z}_{t}^\top(\boldsymbol{x}_{t}^\top\otimes\boldsymbol{V})\right)\\
&\cdot\left(\sum_{t=d+1}^{T}(\boldsymbol{x}_{t}^\top\otimes\boldsymbol{V})^\top\boldsymbol{z}_{t}\boldsymbol{z}_{t}^\top(\boldsymbol{x}_{t}^\top\otimes\boldsymbol{V})\right)^{-1}. \\
\end{aligned}
\end{equation}

\subsubsection{Updating the Variable $\boldsymbol{V}$}

With respect to the variable $\boldsymbol{V}$, the optimization problem is given by
\begin{equation}
\boldsymbol{V}:=\argmin_{\boldsymbol{V}}~\frac{1}{2}\sum_{t=d+1}^{T}\|\boldsymbol{y}_{t}-\boldsymbol{W}\boldsymbol{G}(\boldsymbol{x}_{t}^\top\otimes\boldsymbol{V})^\top\boldsymbol{z}_{t}\|_{2}^{2},
\end{equation}
where $\{\boldsymbol{W},\boldsymbol{G},\boldsymbol{X}\}$ are fixed as known variables. To get the closed-from solution, one is required to reformulate the objective function because the involved Kronecker product complicates the formula. According to the property of Kronecker product (see Lemma~\ref{lemma1}), we can rewrite the objective function as follows,
\begin{equation}
\begin{aligned}
f=&\frac{1}{2}\sum_{t=d+1}^{T}\|\boldsymbol{y}_{t}-\boldsymbol{W}\boldsymbol{G}(\boldsymbol{x}_{t}^\top\otimes\boldsymbol{V})^\top\boldsymbol{z}_{t}\|_{2}^{2} \\
=&\frac{1}{2}\sum_{t=d+1}^{T}\|\boldsymbol{y}_{t}-\boldsymbol{W}\boldsymbol{G}((\boldsymbol{x}_{t}\boldsymbol{z}_{t}^\top)\otimes\boldsymbol{I}_{R})\text{vec}(\boldsymbol{V}^\top)\|_{2}^{2}, \\
\end{aligned}
\end{equation}
where $\text{vec}(\cdot)$ denotes the operator of vectorization.

\begin{lemma}\label{lemma1}
For any $\boldsymbol{x}\in\mathbb{R}^{n},\boldsymbol{Y}\in\mathbb{R}^{p\times q},\boldsymbol{z}\in\mathbb{R}^{p}$, there exists
\begin{equation}
\begin{aligned}
(\boldsymbol{x}^\top\otimes\boldsymbol{Y})^\top\boldsymbol{z}=&(\boldsymbol{x}\otimes\boldsymbol{Y}^\top)\boldsymbol{z} \\
=&\text{vec}(\boldsymbol{Y}^\top\boldsymbol{z}\boldsymbol{x}^\top) \\
=&\text{vec}(\boldsymbol{I}_{q}\boldsymbol{Y}^\top(\boldsymbol{z}\boldsymbol{x}^\top)) \\
=&((\boldsymbol{x}\boldsymbol{z}^\top)\otimes\boldsymbol{I}_{q})\text{vec}(\boldsymbol{Y}^\top).
\end{aligned}
\end{equation}
\end{lemma}

\begin{remark}
This lemma stems from one fundamental property of Kronecker product, which is given by
\begin{equation}
\text{vec}(\boldsymbol{A}\boldsymbol{X}\boldsymbol{B})=(\boldsymbol{B}^\top\otimes\boldsymbol{A})\text{vec}(\boldsymbol{X}),
\end{equation}
for any $\boldsymbol{A}\in\mathbb{R}^{m\times m}$, $\boldsymbol{X}\in\mathbb{R}^{m\times n}$, and $\boldsymbol{B}\in\mathbb{R}^{n\times n}$ commensurate from multiplication in that order.
\end{remark}

Therefore, the partial derivative of $f$ with respect to the vectorized variable $\text{vec}(\boldsymbol{V}^\top)$ is given by
\begin{equation}
\begin{aligned}
\frac{\partial f}{\partial\text{vec}(\boldsymbol{V}^\top)}=&-\sum_{t=d+1}^{T}((\boldsymbol{z}_{t}\boldsymbol{x}_{t}^\top)\otimes\boldsymbol{I}_{R})\boldsymbol{G}^\top\boldsymbol{W}^\top\boldsymbol{y}_{t} \\
&+\sum_{t=d+1}^{T}((\boldsymbol{z}_{t}\boldsymbol{x}_{t}^\top)\otimes\boldsymbol{I}_{R})\boldsymbol{G}^\top\boldsymbol{W}^\top \\
&\cdot\boldsymbol{W}\boldsymbol{G}((\boldsymbol{x}_{t}\boldsymbol{z}_{t}^\top)\otimes\boldsymbol{I}_{R})\text{vec}(\boldsymbol{V}^\top). \\
\end{aligned}
\end{equation}

Let $\frac{\partial f}{\partial\text{vec}(\boldsymbol{V}^\top)}=\boldsymbol{0}$, then there exists a system of linear equations:
\begin{equation}\label{linear_eq_vec_v}
\begin{aligned}
&\sum_{t=d+1}^{T}((\boldsymbol{z}_{t}\boldsymbol{x}_{t}^\top)\otimes\boldsymbol{I}_{R})\boldsymbol{G}^\top\boldsymbol{W}^\top\boldsymbol{W}\boldsymbol{G}((\boldsymbol{x}_{t}\boldsymbol{z}_{t}^\top)\otimes\boldsymbol{I}_{R})\text{vec}(\boldsymbol{V}^\top) \\
=&\sum_{t=d+1}^{T}((\boldsymbol{z}_{t}\boldsymbol{x}_{t}^\top)\otimes\boldsymbol{I}_{R})\boldsymbol{G}^\top\boldsymbol{W}^\top\boldsymbol{y}_{t}. \\
\end{aligned}
\end{equation}
This system of linear equations has a least squares solution. However, the potential inverse of the large matrix would cost $\mathcal{O}((dNR)^3)$, involving high computational complexity. To convert this large-scale and sparse problem into an easy-to-solve problem, we utilize Lemma~\ref{lemma2} to establish an equivalent generalized Sylvester equation with respect to the variable $\boldsymbol{V}$.

\begin{lemma}\label{lemma2}
For any $\boldsymbol{x}\in\mathbb{R}^{n},\boldsymbol{Y}\in\mathbb{R}^{q\times n},\boldsymbol{z}\in\mathbb{R}^{p}$, there exists
\begin{equation}
((\boldsymbol{z}\boldsymbol{x}^\top)\otimes\boldsymbol{I}_{q})\text{vec}(\boldsymbol{Y})=\text{vec}(\boldsymbol{Y}\boldsymbol{x}\boldsymbol{z}^\top).
\end{equation}
\end{lemma}

According to the property of Kronecker product as mentioned in Lemma~\ref{lemma2}, Eq.~\eqref{linear_eq_vec_v} is equivalent to the following generalized Sylvester equation:
\begin{equation}
\begin{aligned}
&\sum_{t=d+1}^{T}((\boldsymbol{z}_{t}\boldsymbol{x}_{t}^\top)\otimes\boldsymbol{I}_{R})\boldsymbol{G}^\top\boldsymbol{W}^\top\boldsymbol{W}\boldsymbol{G}(\boldsymbol{x}_{t}^\top\otimes\boldsymbol{V})^\top\boldsymbol{z}_{t} \\
=&\sum_{t=d+1}^{T}((\boldsymbol{z}_{t}\boldsymbol{x}_{t}^\top)\otimes\boldsymbol{I}_{R})\boldsymbol{G}^\top\boldsymbol{W}^\top\boldsymbol{y}_{t}. \\
\end{aligned}
\end{equation}

As can be seen, it yields a generalized Sylvester equation that infers the solution to $\boldsymbol{V}^\top$:
\begin{equation}\label{matrix_eq_v_transpose}
\sum_{t=d+1}^{T}\boldsymbol{P}_{t}\boldsymbol{x}_{t}\boldsymbol{z}_{t}^\top=\sum_{t=d+1}^{T}\boldsymbol{Q}_{t}\boldsymbol{x}_{t}\boldsymbol{z}_{t}^\top,
\end{equation}
where we define two auxiliary variables $\boldsymbol{P}_{t},\boldsymbol{Q}_{t}\in\mathbb{R}^{R\times R}$ such that
\begin{equation*}
\begin{cases}
\text{vec}(\boldsymbol{P}_{t})\triangleq\boldsymbol{G}^\top\boldsymbol{W}^\top\boldsymbol{W}\boldsymbol{G}(\boldsymbol{x}_{t}^\top\otimes\boldsymbol{V})^\top\boldsymbol{z}_{t},\\
\text{vec}(\boldsymbol{Q}_{t})\triangleq\boldsymbol{G}^\top\boldsymbol{W}^\top\boldsymbol{y}_{t}.
\end{cases}
\end{equation*}

With respect to the variable $\boldsymbol{V}$, the first impulse is to take a transpose operation on both left-hand and right-hand sides of Eq.~\eqref{matrix_eq_v_transpose}. Then the resultant equation is given by
\begin{equation}\label{matrix_eq_v}
\sum_{t=d+1}^{T}\boldsymbol{z}_{t}\boldsymbol{x}_{t}^\top\boldsymbol{P}_{t}^\top=\sum_{t=d+1}^{T}\boldsymbol{z}_{t}\boldsymbol{x}_{t}^\top\boldsymbol{Q}_{t}^\top.
\end{equation}

In this case, we can use the conjugate gradient, a classical and efficient numerical approximation algorithm \cite{golub2013matrix}, to solve the generalized Sylvester equation. To solve Eq.~\eqref{matrix_eq_v} through conjugate gradient, we need to define an operator on the left-hand side of the equation as follows,
\begin{equation}
\mathcal{L}_{v}(\boldsymbol{V})\triangleq\text{vec}\left(\sum_{t=d+1}^{T}\boldsymbol{z}_{t}\boldsymbol{x}_{t}^\top\boldsymbol{P}_{t}^\top\right)\in\mathbb{R}^{dNR}.
\end{equation}

Algorithm~\ref{cong_grad} summarizes the estimation procedure for approximating the solution to the variable $\boldsymbol{V}$ in Eq.~\eqref{matrix_eq_v}. The conjugate gradient method allows one to search for the approximated solution to a system of linear equations with a relatively small number of iterations (e.g., 5 or 10). Admittedly, the conjugate gradient methods with a small number of iterations cannot match the least squares solution. Nevertheless, the numerical approximated solution via conjugate gradient is rather accurate \cite{golub2013matrix}.

\begin{algorithm}
\caption{Conjugate gradient for inferring $\boldsymbol{V}$}
\label{cong_grad}
\begin{algorithmic}[1]
 \renewcommand{\algorithmicrequire}{\textbf{Input:}}
 \renewcommand{\algorithmicensure}{\textbf{Output:}}
 \REQUIRE Data pair $\{\boldsymbol{y}_{t},\boldsymbol{z}_{t}\}$, known variables $\{\boldsymbol{W},\boldsymbol{G},\boldsymbol{X}\}$, initialized $\boldsymbol{V}$, and the maximum iteration $\tilde{L}$ (e.g., the default value as $\tilde{L}=5$).
 \ENSURE Estimated $\boldsymbol{V}$.
 \STATE Let $\mathcal{R}_{v}=\boldsymbol{0}$.
 \FOR {$t=d+1$ to $T$}
 \STATE Compute $\boldsymbol{Q}_{t}$ with $\text{vec}(\boldsymbol{Q}_{t})\triangleq\boldsymbol{G}^\top\boldsymbol{W}^\top\boldsymbol{y}_{t}$. \STATE Take $\mathcal{R}_{v}+=\boldsymbol{z}_{t}\boldsymbol{x}_{t}^\top\boldsymbol{Q}_{t}^\top$.
 \ENDFOR
 \STATE Initialize $\boldsymbol{v}_{0}$ by the vectorized $\boldsymbol{V}$.
 \STATE Compute residual vector $\boldsymbol{r}_{0}=\text{vec}(\mathcal{R}_{v})-\mathcal{L}_v(\boldsymbol{V})$, and $\boldsymbol{q}_0=\boldsymbol{r}_0$.
  \FOR {$\ell = 0$ to $\tilde{L}-1$}
  \STATE Convert vector $\boldsymbol{q}_{\ell}$ into matrix $\boldsymbol{Q}_{\ell}$.
  \STATE Compute $\alpha_{\ell}=\frac{\boldsymbol{r}_{\ell}^\top\boldsymbol{r}_{\ell}}{\boldsymbol{q}_{\ell}^\top\mathcal{L}_v(\boldsymbol{Q}_{\ell})}$.
  \STATE Update $\boldsymbol{v}_{\ell+1}=\boldsymbol{v}_{\ell}+\alpha_{\ell}\boldsymbol{q}_{\ell}$.
  \STATE Update $\boldsymbol{r}_{\ell+1}=\boldsymbol{r}_{\ell}-\alpha_{\ell}\mathcal{L}_v(\boldsymbol{Q}_{\ell})$.
  \STATE Compute $\beta_{\ell}=\frac{\boldsymbol{r}_{\ell+1}^\top\boldsymbol{r}_{\ell+1}}{\boldsymbol{r}_{\ell}^\top\boldsymbol{r}_{\ell}}$.
  \STATE Update $\boldsymbol{q}_{\ell+1}=\boldsymbol{r}_{\ell+1}+\beta_{\ell}\boldsymbol{q}_{\ell}$.
  \ENDFOR
  \STATE Convert vector $\boldsymbol{v}_{\tilde{L}}$ into matrix $\boldsymbol{V}$.
\end{algorithmic}
\end{algorithm}

\subsubsection{Updating the Variable $\{\boldsymbol{x}_{t}\}$}

According to the property of Kronecker product, we can rewrite the objective function of the optimization problem in Eq.~\eqref{time_varying_model} as follows,
\begin{equation}
\begin{aligned}
f=&\frac{1}{2}\sum_{t=d+1}^{T}\|\boldsymbol{y}_{t}-\boldsymbol{W}\boldsymbol{G}(\boldsymbol{x}_{t}^\top\otimes\boldsymbol{V})^\top\boldsymbol{z}_{t}\|_{2}^{2}\\
=&\frac{1}{2}\sum_{t=d+1}^{T}\|\boldsymbol{y}_{t}-\boldsymbol{W}\boldsymbol{G}(\boldsymbol{I}_{R}\otimes(\boldsymbol{V}^\top\boldsymbol{z}_{t}))\boldsymbol{x}_{t}\|_{2}^{2}. \\
\end{aligned}
\end{equation}

Therefore, the optimization problem with respect to $\boldsymbol{x}_{t},\forall t$ now becomes
\begin{equation}
\boldsymbol{x}_{t}:=\argmin_{\boldsymbol{x}_{t}}~\frac{1}{2}\|\boldsymbol{y}_{t}-\boldsymbol{W}\boldsymbol{G}(\boldsymbol{I}_{R}\otimes(\boldsymbol{V}^\top\boldsymbol{z}_{t}))\boldsymbol{x}_{t}\|_{2}^{2},
\end{equation}
while $\{\boldsymbol{W},\boldsymbol{G},\boldsymbol{V}\}$ are fixed as known variables.

Then, we can obtain the partial derivative of $f$ with respect to the variable $\boldsymbol{x}_{t}$ (i.e., the $t$th row of the variable $\boldsymbol{X}$), which is given by
\begin{equation}
\begin{aligned}
\frac{\partial f}{\partial\boldsymbol{x}_{t}}=&(\boldsymbol{I}_{R}\otimes(\boldsymbol{V}^\top\boldsymbol{z}_{t}))^\top\boldsymbol{G}^\top\boldsymbol{W}^\top(\boldsymbol{W}\boldsymbol{G}(\boldsymbol{I}_{R}\otimes(\boldsymbol{V}^\top\boldsymbol{z}_{t}))\boldsymbol{x}_{t}-\boldsymbol{y}_{t}) \\
=&(\boldsymbol{I}_{R}\otimes(\boldsymbol{V}^\top\boldsymbol{z}_{t}))^\top\boldsymbol{G}^\top\boldsymbol{W}^\top\boldsymbol{W}\boldsymbol{G}(\boldsymbol{I}_{R}\otimes(\boldsymbol{V}^\top\boldsymbol{z}_{t}))\boldsymbol{x}_{t} \\
&-(\boldsymbol{I}_{R}\otimes(\boldsymbol{V}^\top\boldsymbol{z}_{t}))^\top\boldsymbol{G}^\top\boldsymbol{W}^\top\boldsymbol{y}_{t}, \\
\end{aligned}
\end{equation}
In this case, letting $\frac{\partial f}{\partial\boldsymbol{x}_{t}}=\boldsymbol{0}$ leads to a least squares solution to the variable $\boldsymbol{x}_{t}$:
\begin{equation}\label{least_square_x}
\boldsymbol{x}_{t}=\left(\boldsymbol{W}\boldsymbol{G}(\boldsymbol{I}_{R}\otimes(\boldsymbol{V}^\top\boldsymbol{z}_{t}))\right)^\dagger\boldsymbol{y}_{t}.
\end{equation}

\subsection{Solution Algorithm}

As mentioned above, the variables $\{\boldsymbol{W},\boldsymbol{G},\boldsymbol{X}\}$ can be computed by the closed-form least squares solutions, while the solution to the variable $\boldsymbol{V}$ can be numerically approximated by the conjugate gradient method in an efficient manner. Starting from the initialized variables by using singular value decomposition, then we update these variables in an iterative routine. In the iterative process, the basic idea of alternating minimization is that we fix the remaining variables when updating one of these variables. Since each subproblem is convex and there is a unique solution to each of the coordinate-wise minimization problems, the convergence of our algorithm through the block coordinate minimization can be therefore verified \cite{harris2021time}. 

\begin{algorithm}
\caption{Time-varying reduced-rank VAR}
\label{trvar}
\begin{algorithmic}[1]
 \renewcommand{\algorithmicrequire}{\textbf{Input:}}
 \renewcommand{\algorithmicensure}{\textbf{Output:}}
 \REQUIRE Data pair $\{\boldsymbol{y}_t,\boldsymbol{z}_{t}\}$ constructed by the time series data $\boldsymbol{S}\in\mathbb{R}^{N\times T}$, $d$ as the order of VAR, $R$ as the low rank ($R\leq\min\{N,T-d\}$), and $L$ as the maximum iteration.
 \ENSURE $\{\boldsymbol{W},\,\boldsymbol{G},\,\boldsymbol{V},\,\boldsymbol{X}\}$.
 \STATE Initialize factor matrices $\boldsymbol{W},\boldsymbol{V},\boldsymbol{X}$ by the $R$ left singular vectors of $\boldsymbol{Y},\boldsymbol{Z},\boldsymbol{S}^\top$, respectively.
  \FOR {$\ell = 0$ to $L-1$}
  \STATE Update $\boldsymbol{G}$ by Eq.~\eqref{least_square_G}.
  \STATE Update $\boldsymbol{W}$ by Eq.~\eqref{least_square_w}.
  \STATE Compute $\boldsymbol{V}$ from Eq.~\eqref{matrix_eq_v} with conjugate gradient (see Algorithm~\ref{cong_grad}).
  \FOR {$t=d+1$ to $T$}
  \STATE Update $\boldsymbol{x}_{t}$ by Eq.~\eqref{least_square_x}.
  \ENDFOR
  \ENDFOR
\end{algorithmic}
\end{algorithm}

\section{Experiments}\label{experiment}

In this section, we evaluate our model on an artificial dataset for fluid dynamics and some real-world datasets, including sea surface temperature, USA surface temperature, and NYC taxi trips. We demonstrate the performance of our model on characterizing the underlying data patterns and system's behavior, especially showing the modeling capability and interpretability on time-varying systems. All the experiment results are replicable at \url{https://github.com/xinychen/vars}.

\subsection{Fluid Dynamics}

\begin{figure*}[ht!]
    \centering
    \begin{subfigure}{0.32\linewidth}
    \centering
    \includegraphics[width = 0.95\textwidth]{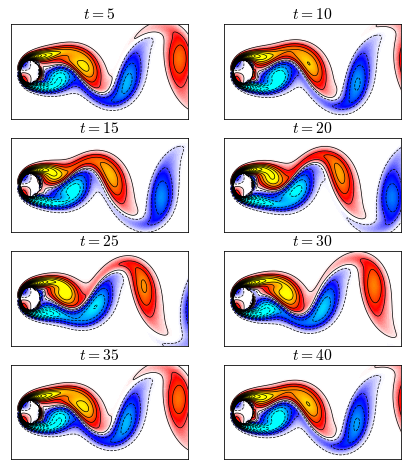}
    \caption{Fluid flow (original data.)}
    \label{fluid_flow_heatmap}
    \end{subfigure}
    \begin{subfigure}{0.32\linewidth}
    \centering
    \includegraphics[width = 0.95\textwidth]{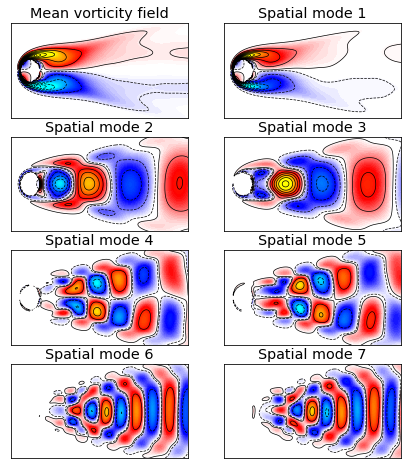}
    \caption{Spatial modes.}
    \label{fluid_flow_spatial_mode}
    \end{subfigure}
    \begin{subfigure}{0.32\linewidth}
    \centering
    \includegraphics[width = 0.95\textwidth]{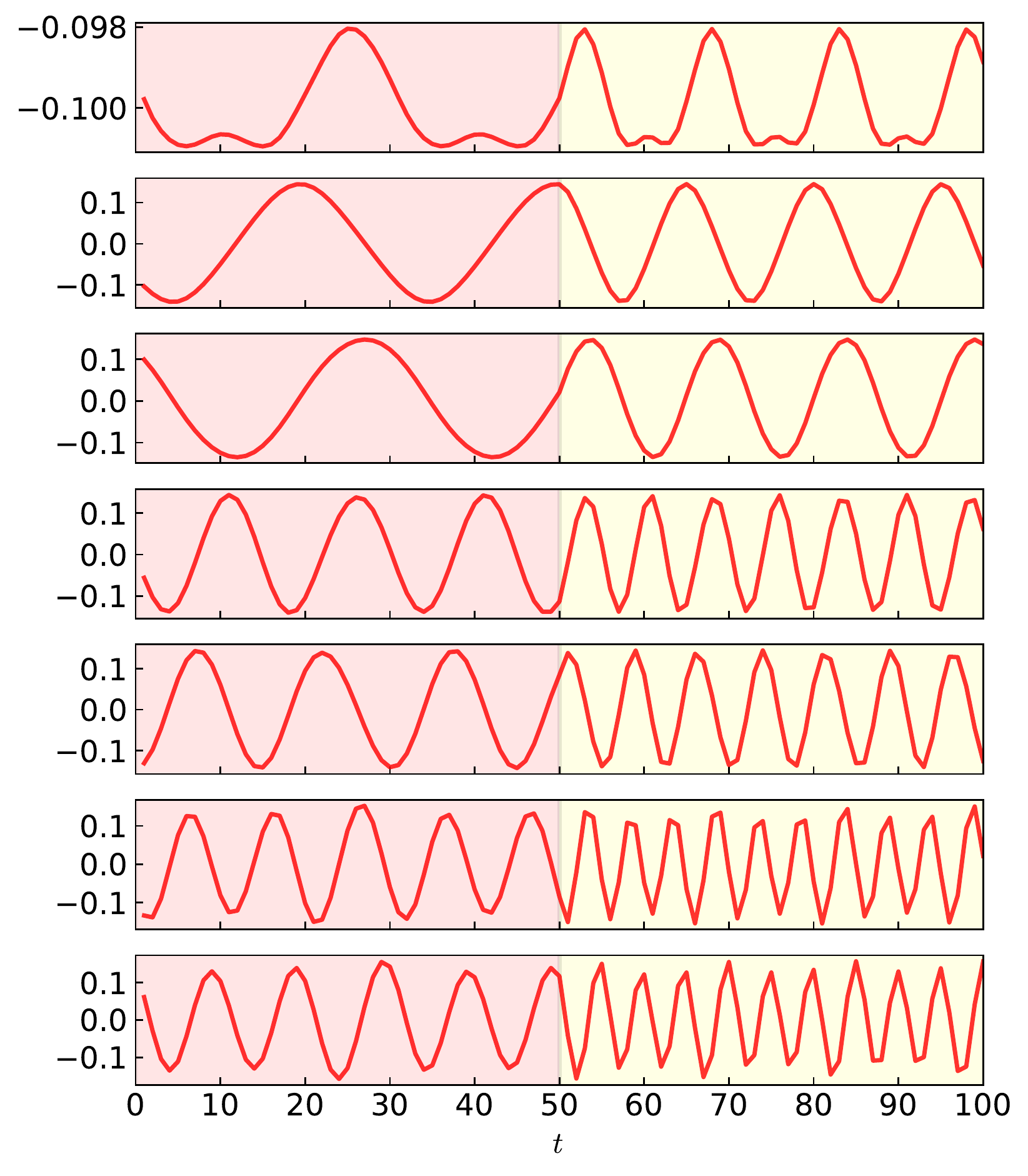}
    \caption{Temporal modes.}
    \label{fluid_temporal_mode}
    \end{subfigure}
    \caption{Fluid flow and spatial/temporal modes to demonstrate the model. (a) Heatmaps (snapshots) of the fluid flow at times $t=5,10,\ldots,40$. It shows that the snapshots at times $t=5$ and $t=35$ are even same, and the snapshots at times $t=10$ and $t=40$ are also even same, allowing one to infer the seasonality as 30 for the first 50 snapshots. (b) Mean vorticity field and spatial modes of the fluid flow. Spatial modes are plotted by the columns of $\boldsymbol{W}$ in which seven panels correspond to the rank $R=7$. Note that the colorbars of all modes are on the same scale. (c) Temporal modes of the fluid flow in $\boldsymbol{X}$. Seven panels correspond to the rank $R=7$.}
\end{figure*}

Investigating fluid dynamic systems is of great interest for uncovering large-scale spatiotemporal coherent structures because dominant patterns exist in the flow field. The data-driven models, such as proper orthogonal decomposition (POD) \cite{berkooz1993proper} and DMD \cite{tu2013dynamic, kutz2016dynamic, wu2021challenges}, have become an important paradigm. To analyze the underlying spatiotemporal patterns of fluid dynamics, we apply our data-driven model to the cylinder wake dataset in which the flow shows a supercritical Hopf bifurcation. The dataset is collected from the fluid flow passing a circular cylinder with laminar vortex shedding at Reynolds number $\mathrm{Re}=100$, which is larger than the critical Reynolds number, using direct numerical simulations of the Navier-Stokes equations.\footnote{\url{http://dmdbook.com/}} This is a representative three-dimensional flow dataset in fluid dynamics, consisting of matrix-variate time series of vorticity field snapshots for the wake behind a cylinder. The dataset is of size $199\times 449\times 150$, representing 199-by-449 vorticity fields with 150 time snapshots (see some examples in Fig.~\ref{fluid_flow_heatmap}).

We first manually build an artificial dataset based on fluid dynamics observations to test the proposed model. First, we can reshape the data as a high-dimensional multivariate time series matrix of size $89351\times 150$. Then, to manually generate a multiresolution system in which the fluid flow takes dynamic-varying system behaviors, we concatenate two parts of data with different frequencies---putting i) the first 50 snapshots (original frequency) together with ii) the uniformly sampled 50 snapshots from the last 100 snapshots (double frequency). As a consequence, the newly-built fluid flow dataset for evaluation has 100 snapshots in total but with a frequency transition in its system behaviors, i.e., possessing different frequencies in two phases. Multiresolution fluid dynamic data come with their own challenges such as multiple frequencies that the standard DMD model cannot work effectively (see Appendix~\ref{appendix_dmd} for further information). Regarding these challenges, finding a spatiotemporal coherent structure from multiresolution data is a challenging task, but of great significance. As demonstrated by \cite{kutz2016multiresolution}, uncovering such multiscale system can recognize and separate multiscale spatiotemporal features.

In our model, rank is a key parameter, which indicates the number of dominant spatial/temporal modes. In practice, a lower rank may only help reveal a few low-frequency dominant modes, but a higher rank would bring some complicated and less interpretable modes, usually referring to high-frequency information \cite{hansen2006deblurring}. This is consistent with the law that nature typically follows---noise is usually dominant at high frequencies and the system signal is more dominant at lower frequencies. On this dataset, we set the rank of our model as $R=7$. Fig.~\ref{fluid_flow_spatial_mode} shows the spatial modes of the fluid flow revealed by our model. It demonstrates that the spatial mode 1 corresponds to a background mode that is not changing over time because it is consistent with the mean vorticity. The other dominant spatial modes essentially show the waves of fluid flow, which look similar to both DMD and POD modes on the same analysis \cite{kutz2016dynamic}. With the increase of rank, the spatial modes can be more detailed. In this case, observing the harmonic frequencies of temporal modes (see Fig.~\ref{fluid_temporal_mode}), the corresponding spatial modes 4 and 5 are more complicated than the spatial modes 2 and 3, while the spatial modes 6 and 7 are more complicated than the spatial modes 4 and 5. Despite recognizing similar spatial modes as POD, our model can also discover the temporal modes that reveal how the system behaviors evolve. Fig.~\ref{fluid_temporal_mode} shows the temporal modes of the fluid flow in $\boldsymbol{X}$. As can be seen, the frequency of all temporal modes in $\boldsymbol{X}$ is changing at the time $t=50$, and all temporal modes can identify the time series with different frequencies of oscillation. The dynamics of fluid flow essentially consist of the phases of two frequencies. Thus, we can emphasize the model's ability for identifying the time-evolving patterns from multiresolution fluid flow. The temporal mode 1 is the most dominant pattern of the fluid flow, corresponding to the spatial mode 1. With higher rank, the frequency of harmonic cycles increases, implying that the importance of the latter modes is secondary and the last spatial/temporal modes represent high-frequency information. Therefore, we can tell that our model can discover both spatial and temporal modes from the spatiotemporal data with time-varying system behaviors.

\begin{figure*}[ht!]
\def\tabularxcolumn#1{m{#1}}
\begin{tabularx}{\textwidth}{@{}cXX@{}}\begin{tabular}{c}
\subfloat[Distribution of mean temperature.]{\includegraphics[width = 0.35\textwidth]{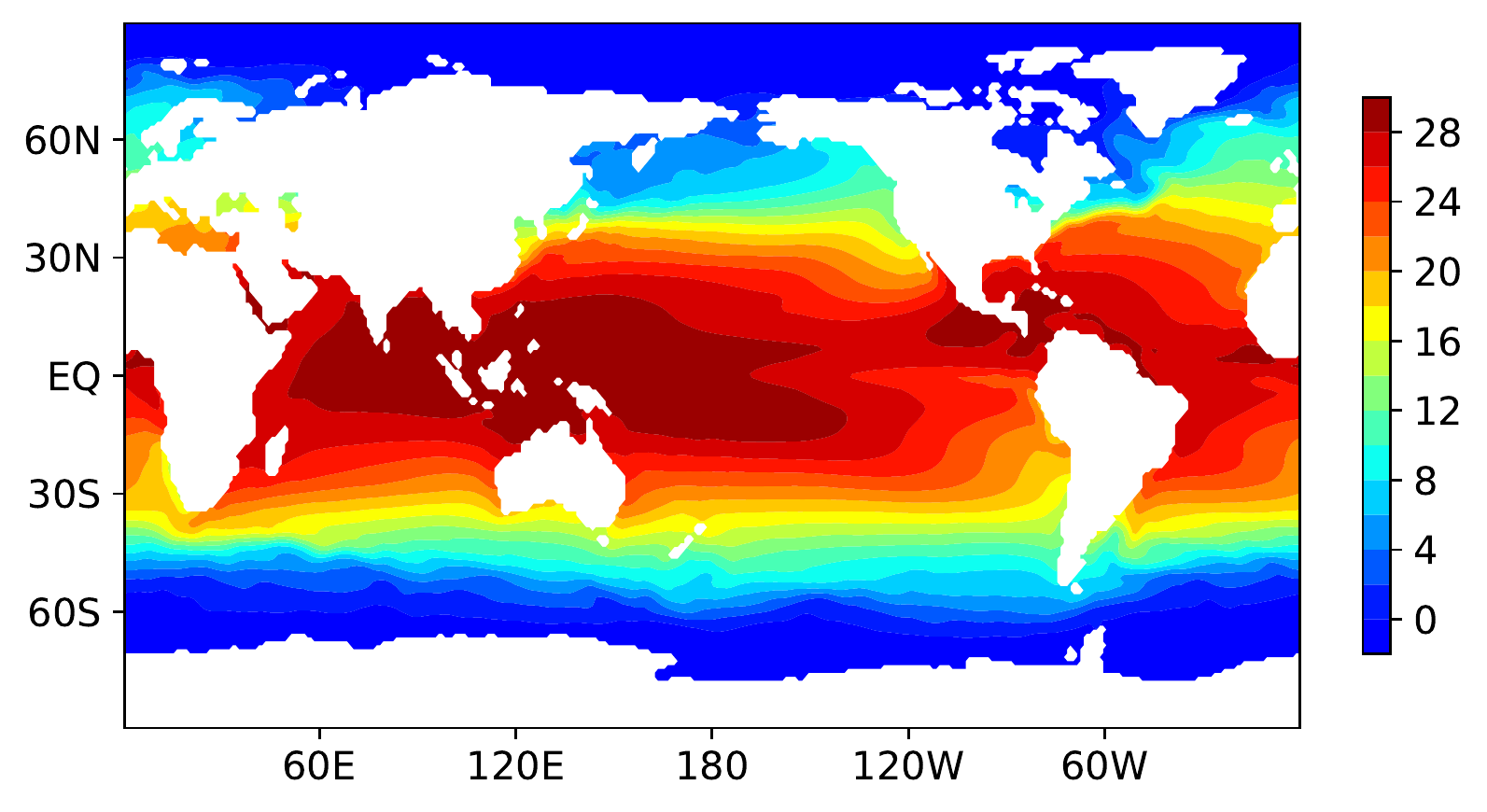}\label{mean_temperature}}\\
\subfloat[Time series of mean temperature.]{\includegraphics[width = 0.45\textwidth]{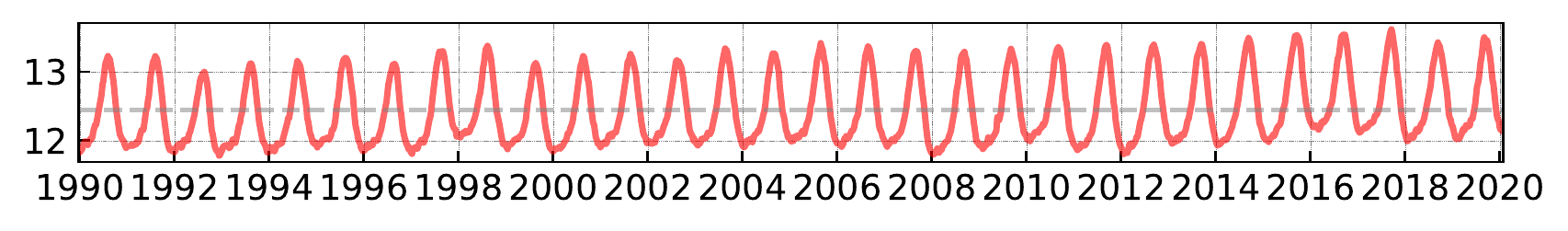}\label{mean_temperature_time_series}} \\
\subfloat[Temporal modes.]{\includegraphics[width = 0.45\textwidth]{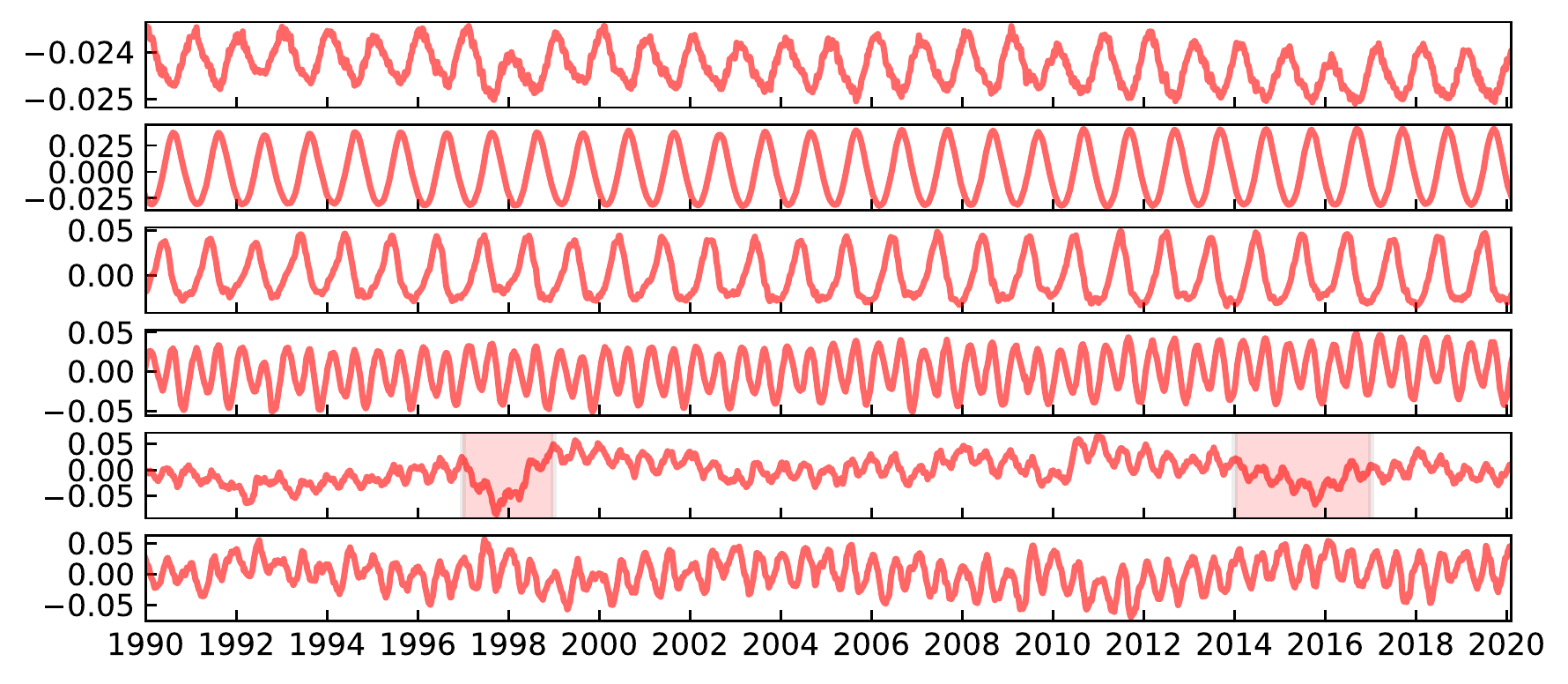}\label{temperature_temporal_mode}}\end{tabular}&
\subfloat[Spatial modes.]{\includegraphics[width = 0.5\textwidth]{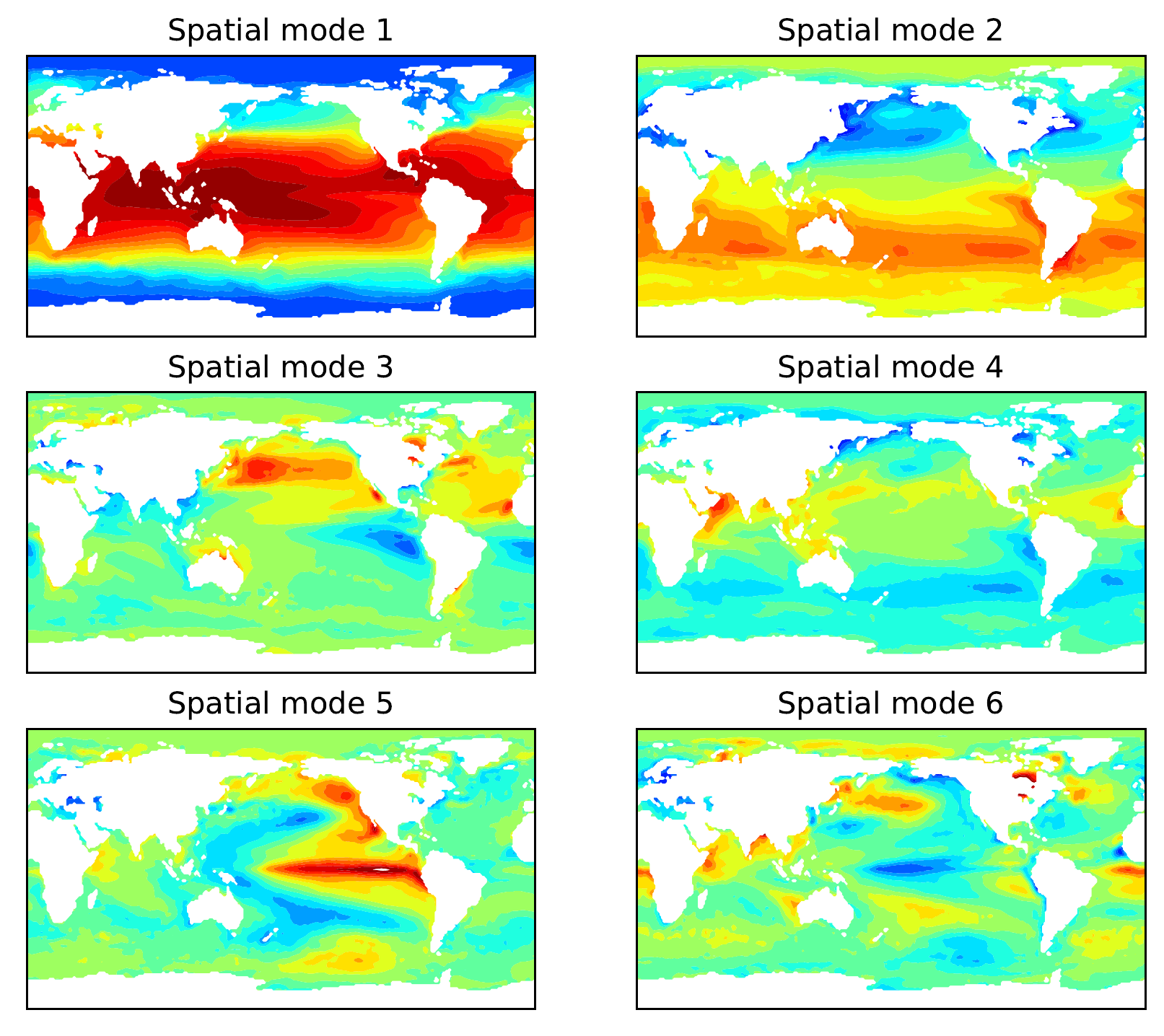}\label{temperature_mode_trvar}}
\end{tabularx}
\caption{Model application and results on SST data from 1990 to 2019. (a) Geographical distribution of long-term mean values of the SST over the 30-year period. (b) The time series of mean values of the SST over the 30-year period. The whole average temperature is 12.45$^\circ\text{C}$. (c) Temporal modes of the SST achieved by our model. From the top panel to the bottom panel, we have the temporal modes 1, \ldots, 6, respectively. (d) Geographical distribution of spatial modes of the SST achieved by our model. Spatial modes are plotted by the columns of $\boldsymbol{W}$ in which six panels correspond to the rank $R=6$.}
\end{figure*}

\subsection{Sea Surface Temperature (SST)}

The oceans play a very important role in the global climate system. Exploiting the SST data allows one to monitor the climate change and understand the dynamical processes of energy exchange at the sea surface \cite{deser2010sea, kutz2016multiresolution}. Here, we consider the SST dataset that covers weekly means of temperature on the spatial resolution of (1 degree latitude, 1 degree longitude)-grid, and there are $180\times 360$ global grids (i.e., 64,800 grids) in total.\footnote{\url{https://psl.noaa.gov/data/gridded/data.noaa.oisst.v2.html}} The dataset spans a 30-year period from 1990 to 2019, and the time dimension is of length 1,565 (weeks). Therefore, the data can be represented as a matrix of size $64800\times 1565$, which seems to be high-dimensional. Fig.~\ref{mean_temperature} exhibits the long-term mean values of the SST dataset over the 30-year period. It is not hard to see the most basic patterns of geographical distribution of SST. Fig.~\ref{mean_temperature_time_series} shows the yearly cycle ($\approx52$ weeks) of time series of the mean temperature.

Using the proposed model with rank $R=6$, we plot the temporal modes and spatial modes of the SST data in Fig.~\ref{temperature_temporal_mode} and \ref{temperature_mode_trvar}, respectively. Our model can identify both quasi-periodic and non-periodic behaviors and patterns from complicated and noisy signals. In contrast, DMD models are empirically demonstrated to be sensitive to the data noise in SST \cite{kutz2016dynamic}. The temporal modes 1, 2, 3, and 4 take a yearly cycle/rhythm, showing dominant patterns of SST. Notably, the spatial modes revealed by our data-driven model is consistent with the previous literature \cite{kutz2016multiresolution}. Specifically, the spatial mode 1, as a background mode, is consistent with the long-term mean temperature. The other spatial modes explain the deviations from the long-term monthly means respectively. It is worth noting that some special oceanic events can also be revealed in these spatial and temporal modes. For instance, the spatial/temporal modes 3 and 4 convey the two Southern Annular Modes in Southern Ocean as well as the positive and negative phases of the North Atlantic Oscillation. In addition, the spatial/temporal modes 5 and 6 demonstrate the phenomenon of El Ni\~{n}o Southern Oscillation (with both of El Ni\~{n}o and La Ni\~{n}a), as well as the Pacific Decadal Oscillation. Combining with the temporal mode 5 in Fig.~\ref{temperature_temporal_mode}, we can directly identify two strongest El Ni\~{n}o events on record, i.e., happened on 1997--98 and 2014--16, corresponding to the spatial mode 5 as highlighted in Fig.~\ref{temperature_mode_trvar}. These results can also be generated through the multiresolution DMD model \cite{kutz2016multiresolution}, but the multiresolution DMD are incapable of discovering the complex SST systems.

\subsection{USA Surface Temperature Data}

\begin{figure}[ht]
    \centering
    \begin{subfigure}{1\linewidth}
    \centering
    \includegraphics[width = 0.65\textwidth]{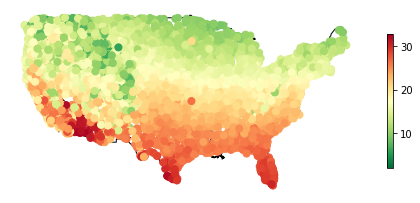}
    \caption{Distribution of mean temperature.}
    \label{usa_temp_spatial_dist}
    \end{subfigure}
    \begin{subfigure}{0.9\linewidth}
    \centering
    \includegraphics[width = 0.9\textwidth]{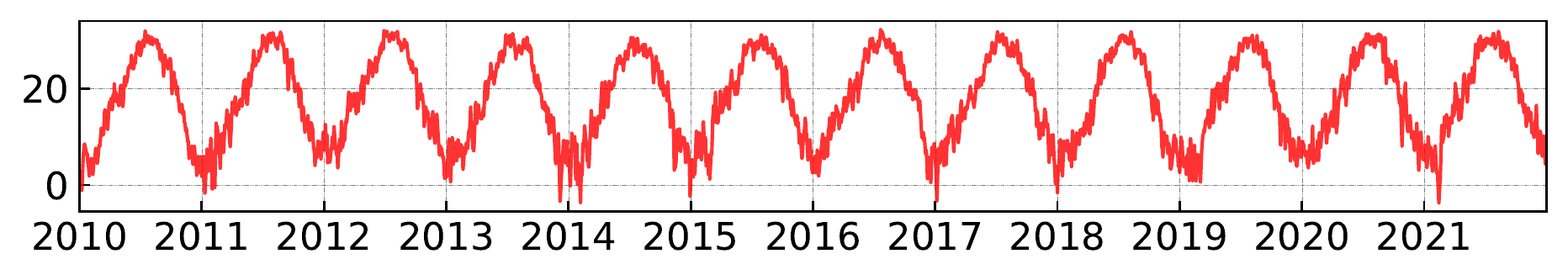}
    \caption{Time series of mean temperature.}
    \label{usa_temp_time_series}
    \end{subfigure}
    \caption{Mean temperature of the maximum daily temperature data in the United States Mainland from 2010 to 2021. (a) Geographical distribution of the long-term mean temperature over the 12-year period. (b) The time series of mean temperature over the 12-year period.}
\end{figure}

Daymet project provides long-term and continuous estimates of daily weather parameters such as maximum and minimum daily temperature for North America.\footnote{\url{https://daac.ornl.gov/DAYMET}} There are 5,380 stations over the United States Mainland. In this work, we use the daily maximum temperature data in the United States Mainland from 2010 to 2021 (i.e., 12 years or 4,380 days in total) for evaluation. The data can be represented as a matrix of size $5380\times 4380$. In particular, we apply the nonstationary temporal matrix factorization model \cite{chen2022nonstationary} to impute 2.67\% missing values in the original data and conduct the following evaluation on the recovered data. Fig.~\ref{usa_temp_spatial_dist} shows the long-term mean values of the temperature dataset over the 12-year period, demonstrating the most basic patterns of geographical distribution of the temperature data. Fig.~\ref{usa_temp_time_series} demonstrates the mean temperature changing over time, showing strong yearly rhythms.

\begin{figure*}[ht!]
    \centering
    \begin{subfigure}{0.9\linewidth}
    \centering
    \includegraphics[width = 1\textwidth]{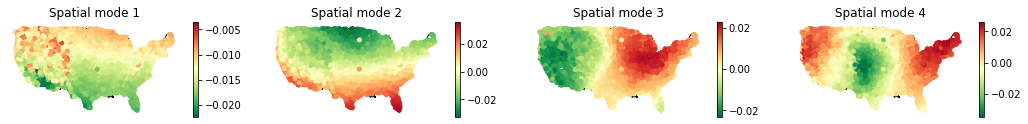}
    \caption{Spatial modes.}
    \label{usa_temp_spatial_modes}
    \end{subfigure}
    \begin{subfigure}{0.45\linewidth}
    \centering
    \includegraphics[width = 1\textwidth]{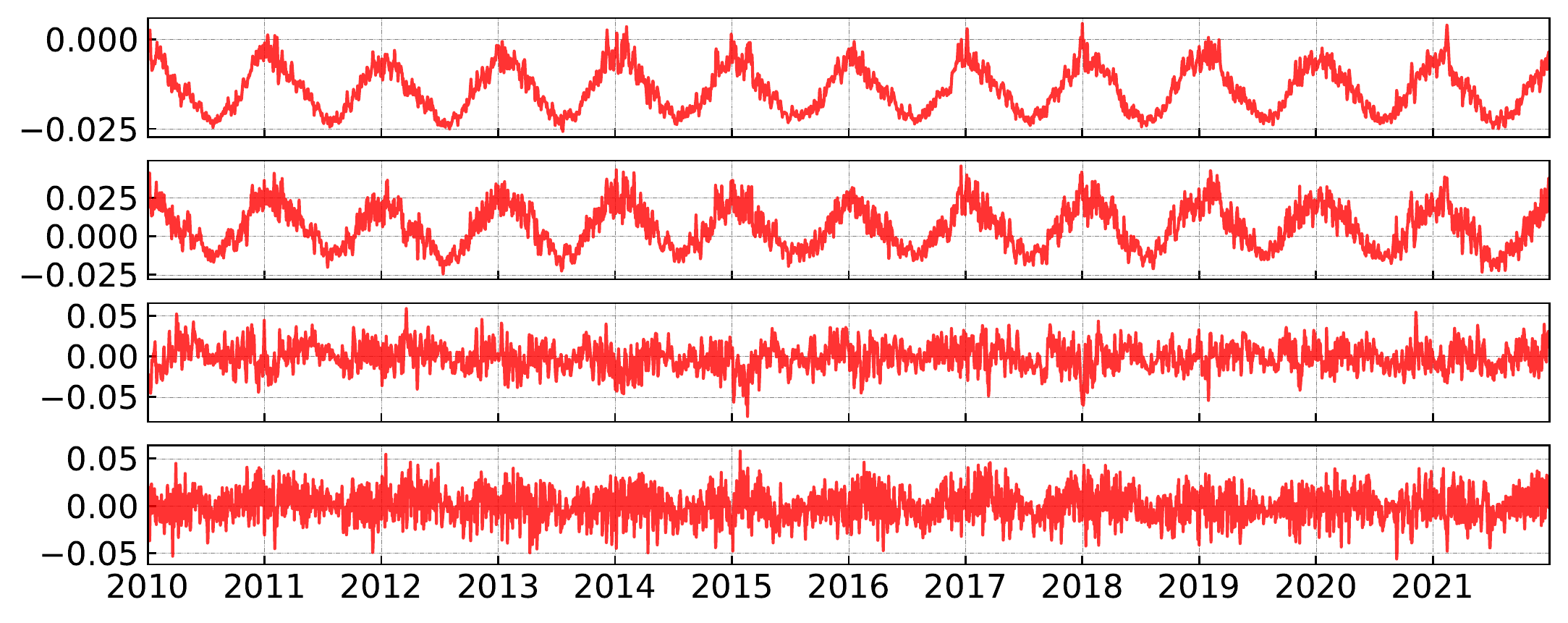}
    \caption{Temporal modes over the 12-year period.}
    \label{usa_temp_temporal_modes}
    \end{subfigure}
    \begin{subfigure}{0.45\linewidth}
    \centering
    \includegraphics[width = 1\textwidth]{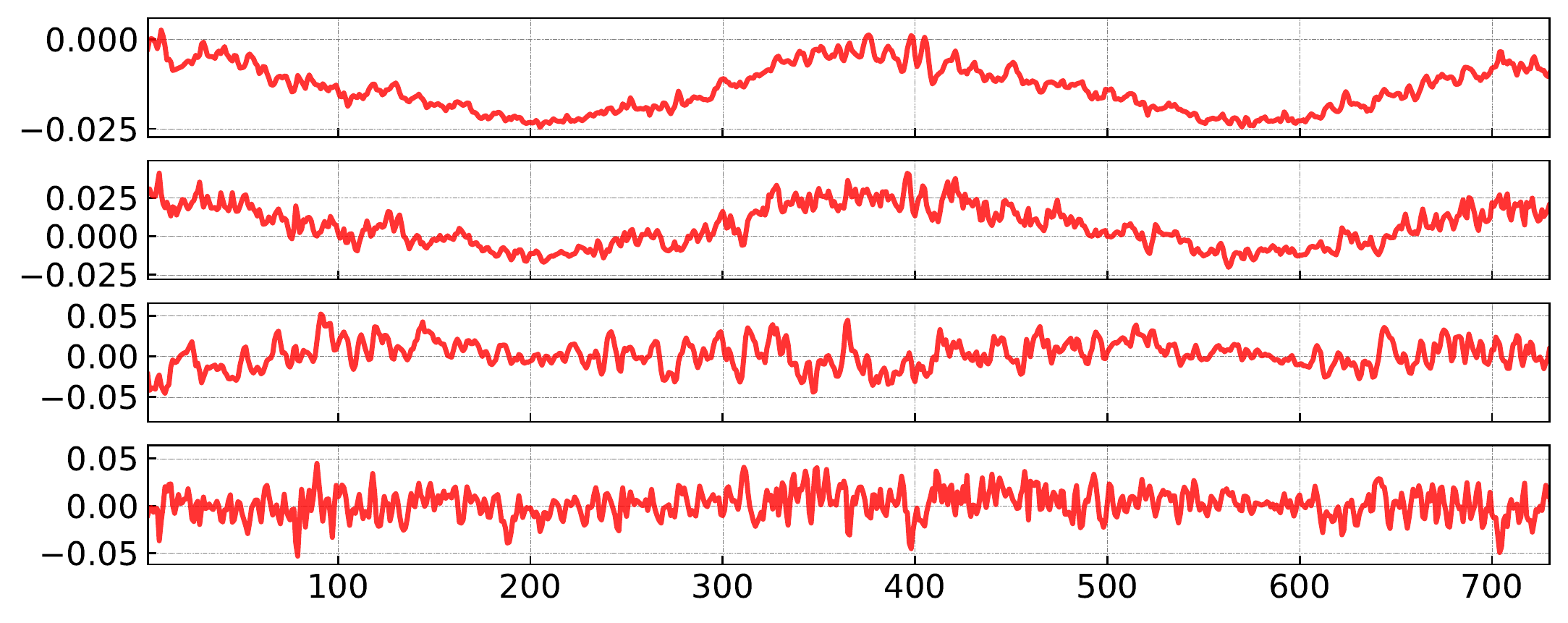}
    \caption{Temporal modes over the two-year period.}
    \label{usa_temp_temporal_modes_zoom_in}
    \end{subfigure}
    \caption{Model application and results on the daily temperature data in the United States Mainland. (a) Geographical distribution of four spatial modes of the temperature data achieved by our model. (b) Four temporal modes (over the 12-year period) of the temperature data achieved by our model. From the top panel to the bottom panel, we have the temporal modes 1-4, respectively. (c) Four temporal modes (during the two years (730 days) from 2010 to 2011) of the temperature data achieved by our model.}
\end{figure*}

Fig.~\ref{usa_temp_spatial_modes} shows the geographical distribution of the spatial modes of the temperature data revealed by $\boldsymbol{W}$ in our model, while Fig.~\ref{usa_temp_temporal_modes} and \ref{usa_temp_temporal_modes_zoom_in} visualize the temporal modes of the temperature data revealed by $\boldsymbol{X}$ in our model. The spatial mode 1 demonstrates the most dominant mode which is consistent with the mean temperature as shown in Fig.~\ref{usa_temp_spatial_dist}. In the meanwhile, the temporal mode 1 shows strong yearly cycles/rhythms underlying the daily temperature. The temporal mode 2 shows the similar time series curve as the temporal mode 1, but their values are quite different. The spatial mode 2 highlights the relatively hot areas and relatively cold areas. The temporal modes 3 and 4 are rather complicated, but the corresponding spatial modes 3 and 4 are intuitive for understanding the geographical distribution of these patterns. The spatial mode 3 roughly highlights the eastern areas and the western areas, identifying different characteristics. The spatial mode 4 identifies three areas in which the eastern and western areas follow the similar feature and the central areas take another feature.

\begin{figure*}[ht!]
    \centering
    \begin{subfigure}{0.9\linewidth}
    \centering
    \includegraphics[width = 1\textwidth]{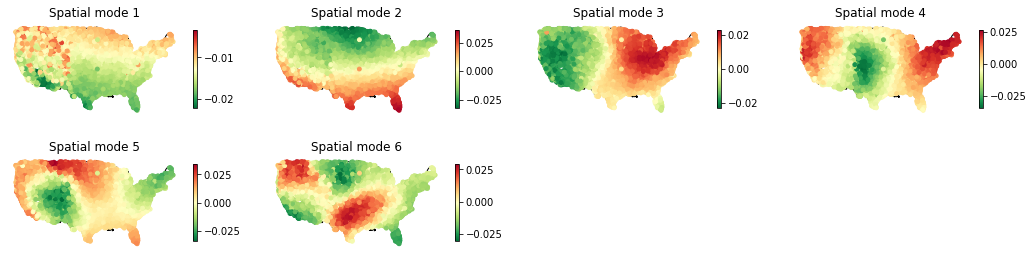}
    \caption{Rank $R=6$.}
    \label{usa_temp_spatial_modes_rank_6}
    \end{subfigure}
    \begin{subfigure}{0.9\linewidth}
    \centering
    \includegraphics[width = 1\textwidth]{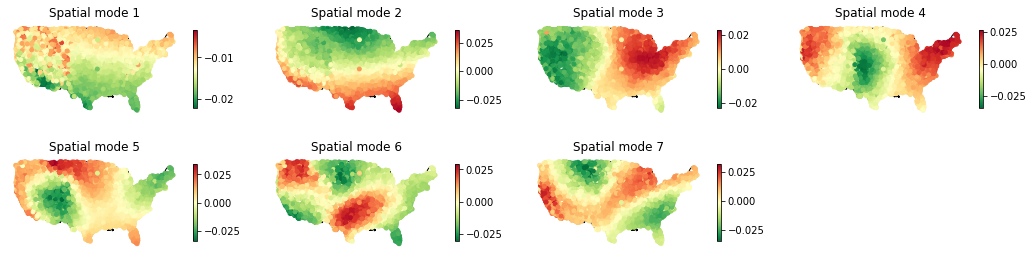}
    \caption{Rank $R=7$.}
    \label{usa_temp_spatial_modes_rank_7}
    \end{subfigure}
    \begin{subfigure}{0.9\linewidth}
    \centering
    \includegraphics[width = 1\textwidth]{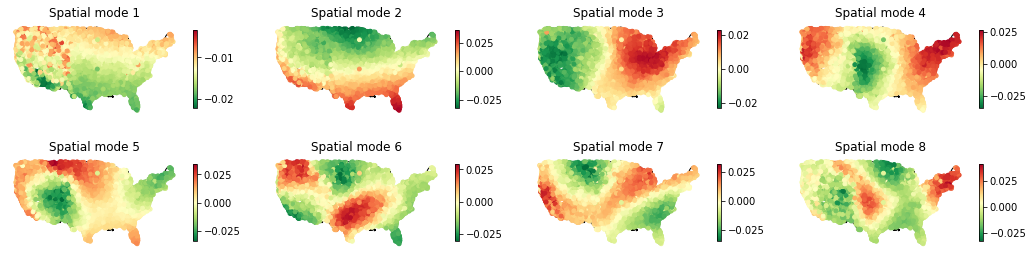}
    \caption{Rank $R=8$.}
    \label{usa_temp_spatial_modes_rank_8}
    \end{subfigure}
    \caption{Geographical distribution of spatial modes of the temperature data achieved by our model.}
\end{figure*}

Recall that rank is a key parameter of our model, which also refers to the number of spatial/temporal modes. Thus, the prescribed rank determines how many dominant modes we discover. Fig.~\ref{usa_temp_spatial_modes_rank_6}, \ref{usa_temp_spatial_modes_rank_7}, and \ref{usa_temp_spatial_modes_rank_8} show the geographical distribution of spatial modes of the temperature data achieved by our model with ranks $R=6,7,8$, respectively. It is clear to see that the first four spatial modes achieved by our model with ranks $R=6,7,8$ are same as the spatial modes in Fig.~\ref{usa_temp_spatial_modes}. As shown in Fig.~\ref{usa_temp_spatial_modes_rank_6}, \ref{usa_temp_spatial_modes_rank_7}, and \ref{usa_temp_spatial_modes_rank_8}, the spatial modes achieved by our model with a relatively large rank (e.g., $R=8$) can cover the the spatial modes achieved by our model with a relatively small rank (e.g., $R=6,7$). Furthermore, as the rank increases, the spatial modes achieved by our model tend to reveal higher-frequency information and more complicated patterns than the model with a relatively small rank. Thus, these findings demonstrate that our model is capable of effectively discovering dominant spatial and temporal patterns from the real-world data.

\begin{figure*}[ht!]
    \centering
    \begin{subfigure}{0.9\linewidth}
    \centering
    \includegraphics[width = 1\textwidth]{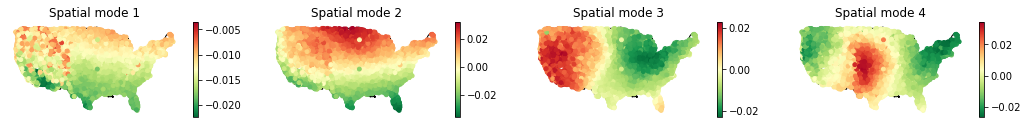}
    \caption{Order $d=2$.}
    \label{usa_temp_spatial_modes_rank_4_d2}
    \end{subfigure}
    \begin{subfigure}{0.9\linewidth}
    \centering
    \includegraphics[width = 1\textwidth]{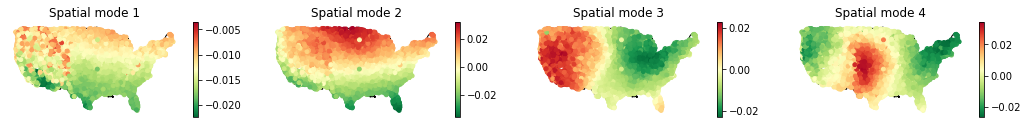}
    \caption{Order $d=3$.}
    \label{usa_temp_spatial_modes_rank_4_d3}
    \end{subfigure}
    \caption{Geographical distribution of four spatial modes of the temperature data achieved by our model with rank $R=4$.}
\end{figure*}

In the above experiments, we use the model with order $d=1$. In practice, we can also set a relatively large order for the time-varying VAR. Recall that we have two variables associated with the spatial dimension, i.e., $\boldsymbol{W}\in\mathbb{R}^{N\times R}$ and $\boldsymbol{V}\in\mathbb{R}^{(dN)\times R}$, in which we assume that the spatial modes revealed by $\boldsymbol{W}$ do not change with the increase of the order $d$. Fig.~\ref{usa_temp_spatial_modes_rank_4_d2} and \ref{usa_temp_spatial_modes_rank_4_d3} show the spatial modes achieved by our model with rank $R=4$ and order $d=2,3$. These results demonstrate that our model discovers similar spatial modes with different orders.

\subsection{NYC Taxi Trips}

\begin{figure*}[ht!]
    \centering
    \begin{subfigure}{0.9\linewidth}
    \centering
    \includegraphics[width = 1\textwidth]{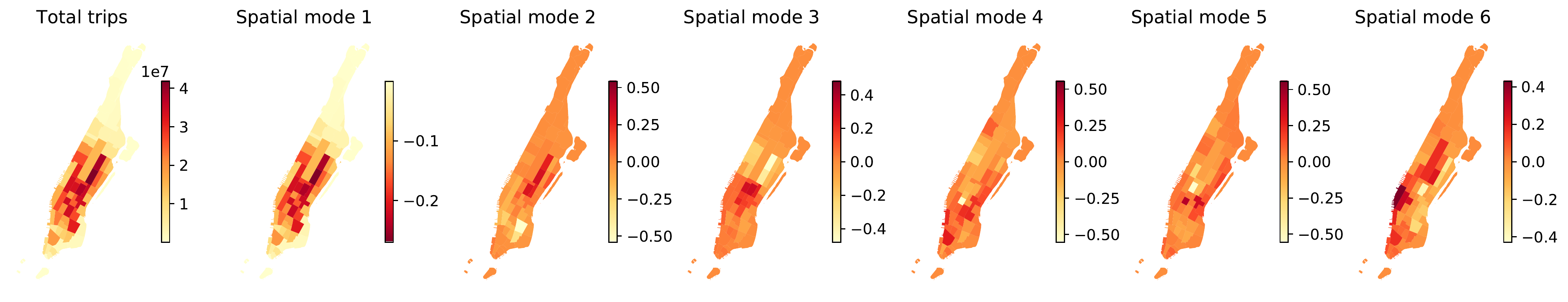}
    \caption{Total pickup trips and spatial modes.}
    \label{pickup_trips}
    \end{subfigure}
    \begin{subfigure}{0.3\linewidth}
    \centering
    \includegraphics[width = 1\textwidth]{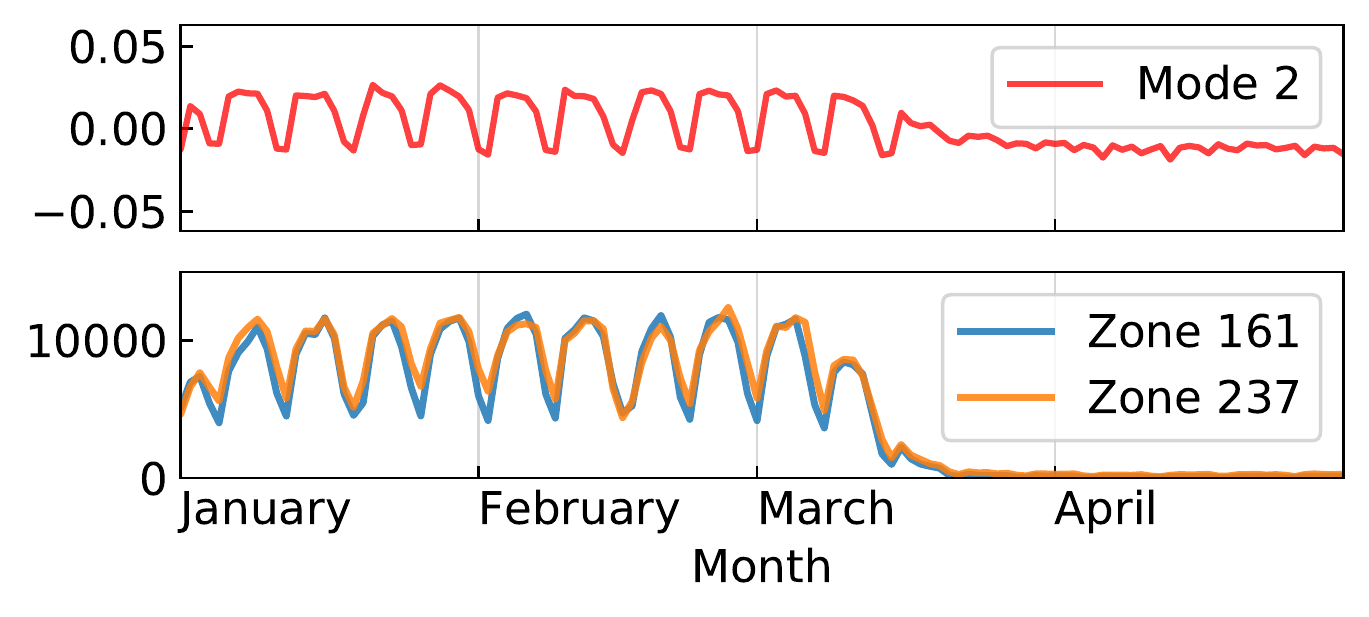}
    \caption{Temporal mode 2 and taxi trips.}
    \label{taxi_temporal_mode_2_pickup}
    \end{subfigure}
    \begin{subfigure}{0.3\linewidth}
    \centering
    \includegraphics[width = 1\textwidth]{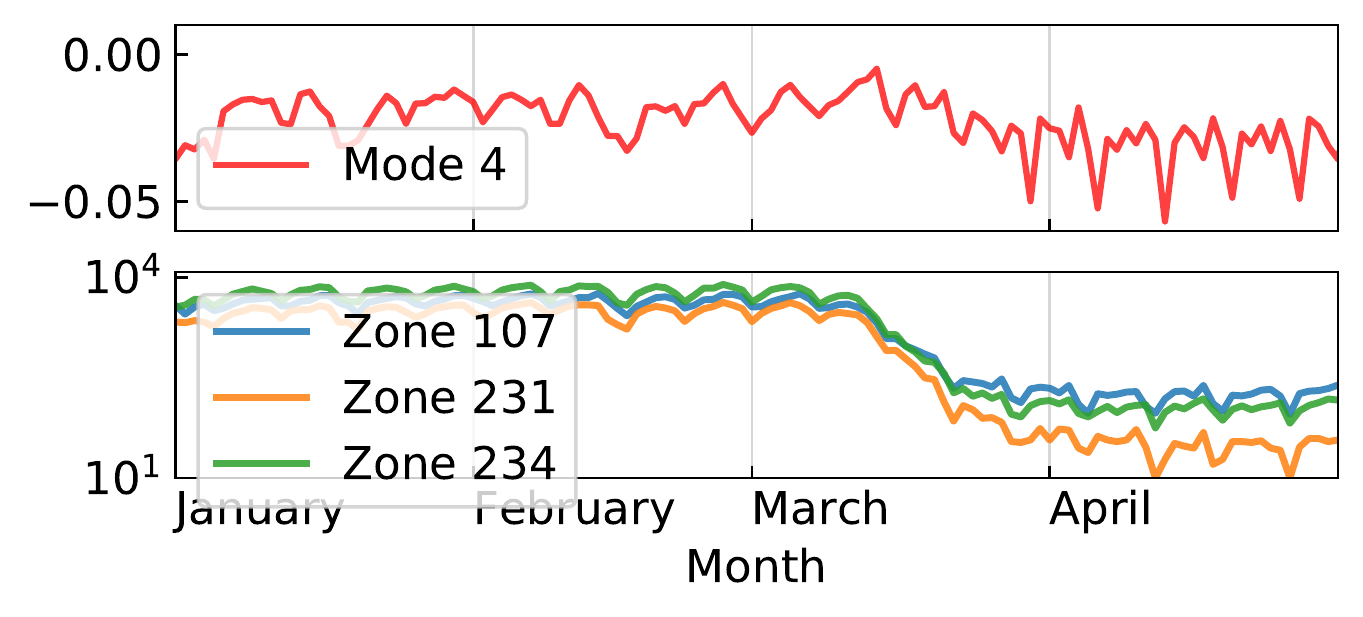}
    \caption{Temporal mode 4 and taxi trips.}
    \label{taxi_temporal_mode_4_pickup}
    \end{subfigure}
    \begin{subfigure}{0.3\linewidth}
    \centering
    \includegraphics[width = 1\textwidth]{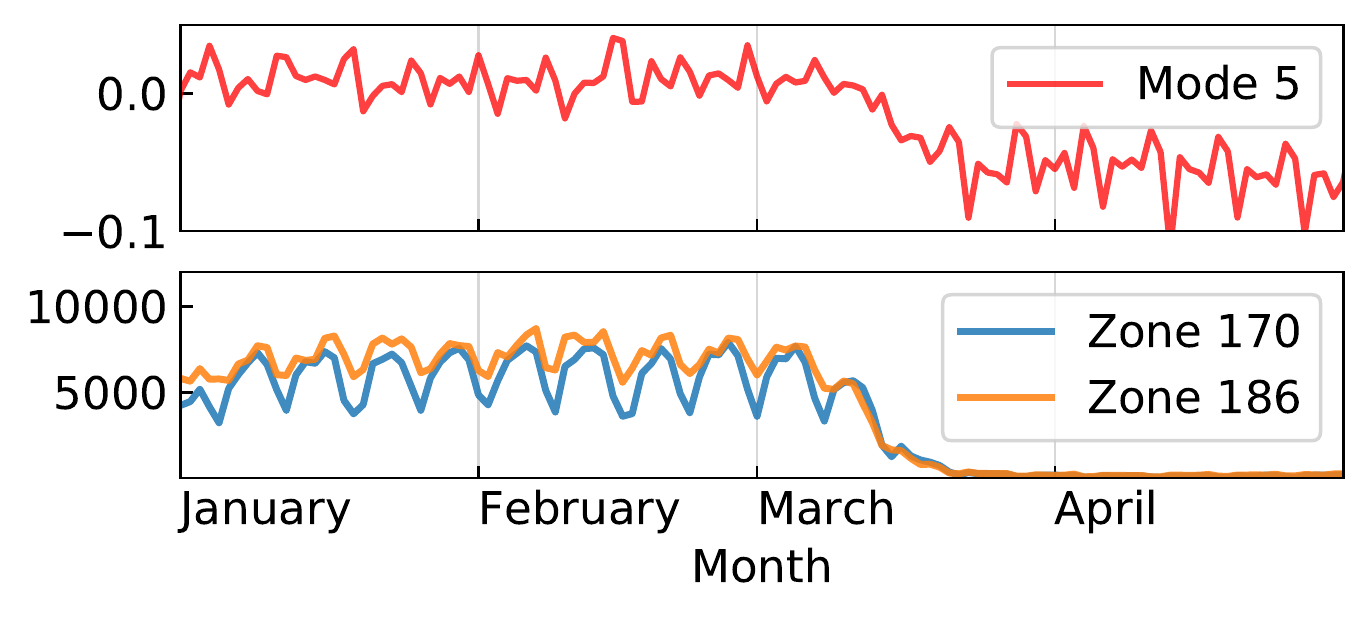}
    \caption{Temporal mode 5 and taxi trips.}
    \label{taxi_temporal_mode_5_pickup}
    \end{subfigure}
    \caption{NYC taxi pickup trips and their spatial and temporal modes achieved by our model. We zoom in the temporal modes in the first four months of 2020. These modes reveal the total traffic reduction due to the COVID-19 pandemic since March 2020. (a) Total trips and spatial modes revealed by $\boldsymbol{W}$. (b-d) refer to temporal mode 2, 4, 5, respectively; note that the bottom panels of these temporal modes uncover the trip time series of certain taxi zones.}
\end{figure*}

\begin{figure*}[ht!]
    \centering
    \begin{subfigure}{0.9\linewidth}
    \centering
    \includegraphics[width = 1\textwidth]{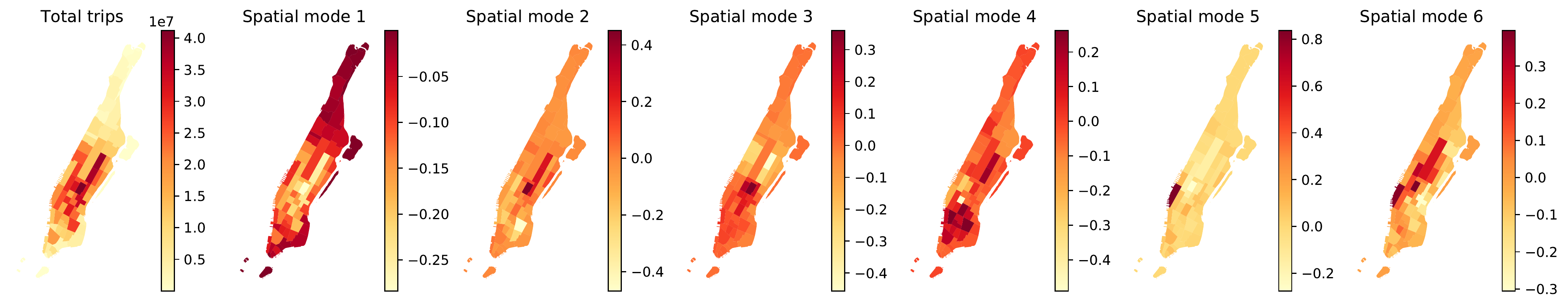}
    \caption{Total dropoff trips and spatial modes.}
    \label{dropoff_trips}
    \end{subfigure}
    \begin{subfigure}{0.3\linewidth}
    \centering
    \includegraphics[width = 1\textwidth]{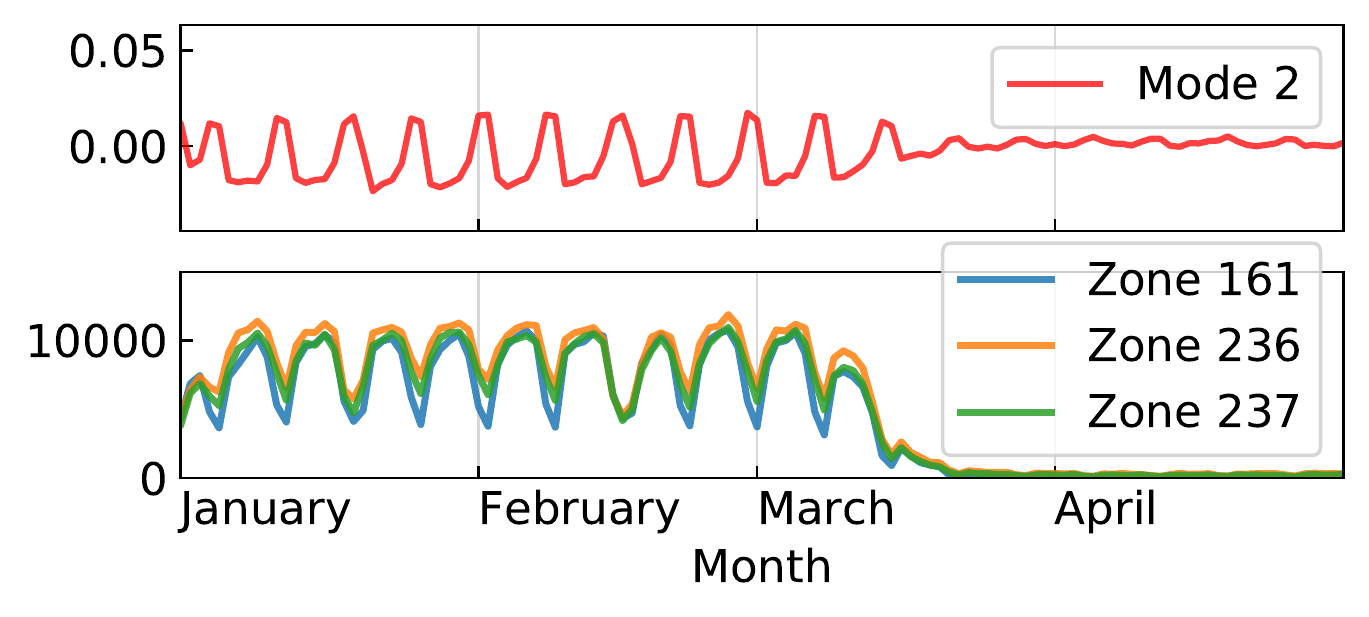}
    \caption{Temporal mode 2 and taxi trips.}
    \label{taxi_temporal_mode_2_dropoff}
    \end{subfigure}
    \begin{subfigure}{0.3\linewidth}
    \centering
    \includegraphics[width = 1\textwidth]{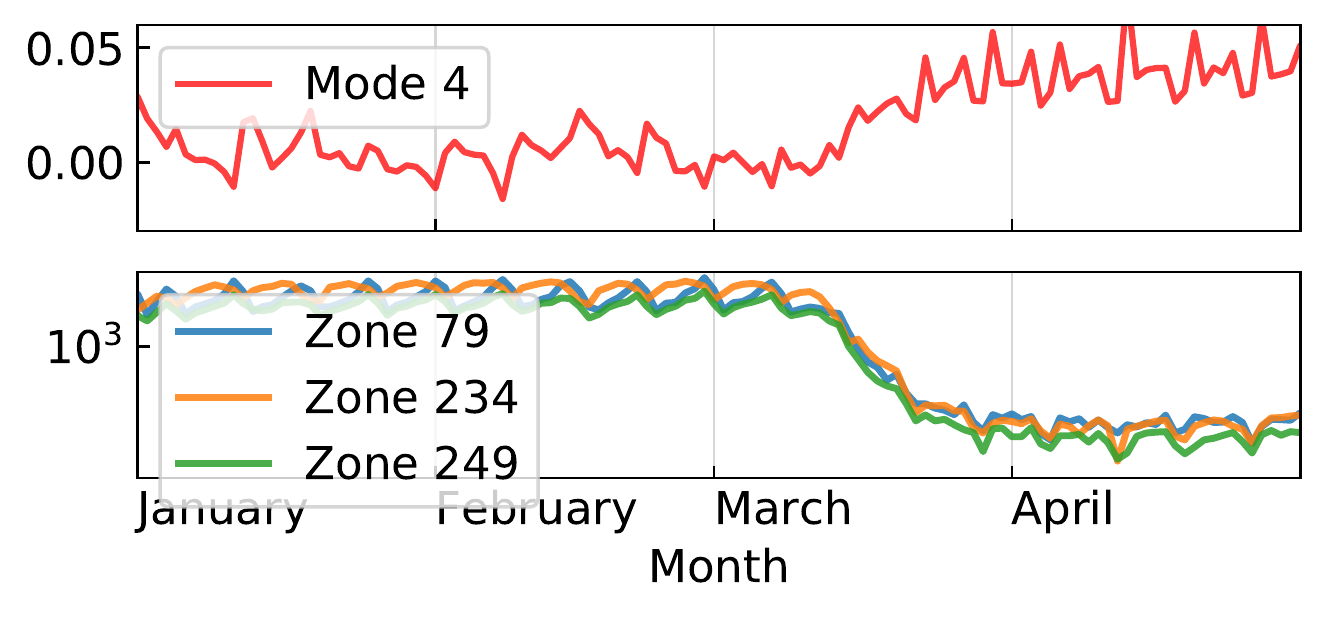}
    \caption{Temporal mode 4 and taxi trips.}
    \label{taxi_temporal_mode_4_dropoff}
    \end{subfigure}
    \begin{subfigure}{0.3\linewidth}
    \centering
    \includegraphics[width = 1\textwidth]{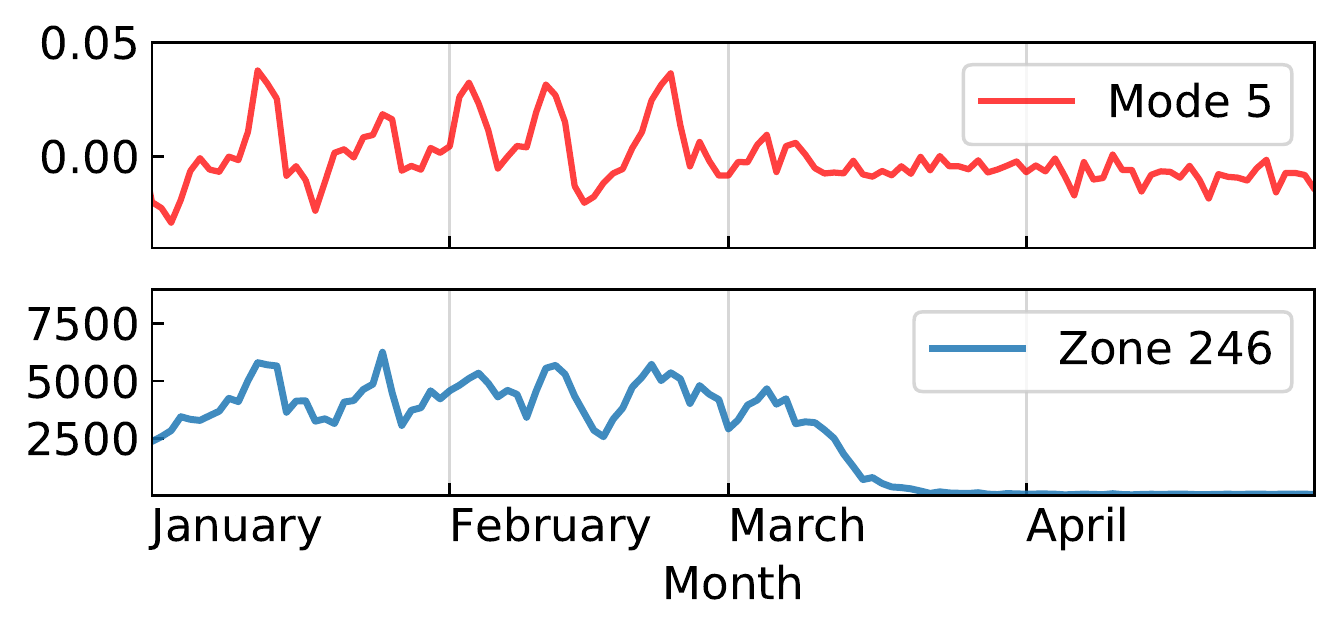}
    \caption{Temporal mode 5 and taxi trips.}
    \label{taxi_temporal_mode_5_dropoff}
    \end{subfigure}
    \caption{NYC taxi dropoff trips and their spatial and temporal modes achieved by our model. We zoom in the temporal modes in the first four months of 2020. These modes reveal the total traffic reduction due to the COVID-19 pandemic since March 2020. (a) Total trips and spatial modes revealed by $\boldsymbol{W}$. (b-d) refer to temporal mode 2, 4, 5, respectively; note that the bottom panels of these temporal modes uncover the trip time series of certain taxi zones.}
\end{figure*}

Human mobility data are always associated with strong quasi-seasonality (e.g., weekly rhythms) and trends (e.g., decline of trips due to unplanned events like COVID-19), it is challenging to characterize these nonlinear and time-varying system behaviors. In this work, we consider to use a NYC (yellow) taxi trip dataset.\footnote{\url{https://www1.nyc.gov/site/tlc/about/tlc-trip-record-data.page}} We use 69 zones in Manhattan as pickup/dropoff zones and aggregate daily taxi trip volume of the data from 2012 to 2021. Therefore, the daily trip volume tensor is of size $69\times 69\times 3653$. In the following analysis, we aggregate the the trip volume tensor as a pickup trip volume matrix and a dropoff trip volume matrix, both of size $69\times 3653$. The left panel of Fig.~\ref{pickup_trips} and \ref{dropoff_trips} shows the total pickup trips and total dropoff trips, respectively. As can be seen, most pickup/dropoff trips are created in the central urban areas of Manhattan.

\begin{figure}[ht!]
    \centering
    \includegraphics[width = 0.2\textwidth]{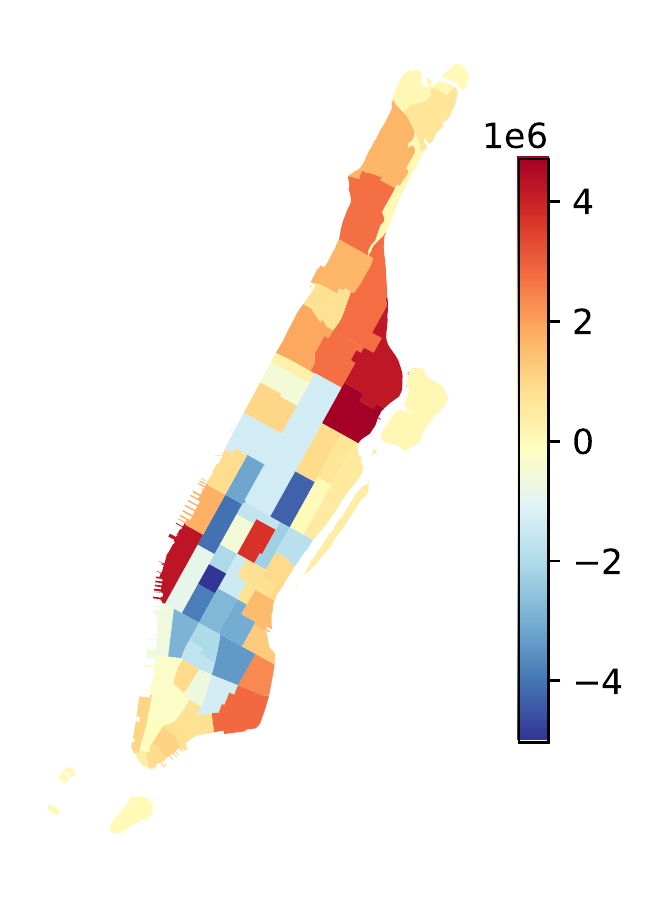}
    \caption{Total dropoff trips minus total pickup trips in the 69 zones of Manhattan.}
    \label{taxi_dropoff_minus_pickup}
\end{figure}

Fig.~\ref{taxi_dropoff_minus_pickup} visualizes the long-term changes of human mobility via the dropoff trips minus the pickup trips. The dropoff trips are greater than the pickup trips in edges of urban areas (see the red zones such as the Upper East Side of Manhattan). In contrast, the pickup trips are greater than the dropoff trips in the central urban areas (see the blue zones). Therefore, it demonstrates the difference between the pickup trips and the dropoff trips in the spatial context.

As shown in Fig.~\ref{pickup_trips} and \ref{dropoff_trips}, the spatial and temporal modes can explain the long-term trip patterns of NYC taxi trips. The spatial mode 1 essentially demonstrates the long-term patterns, which is consistent with the total trips (see the left panel of Fig.~\ref{pickup_trips} and \ref{dropoff_trips}). Other spatial modes can be used to reveal the trip patterns of some specific zones. In terms of pickup trips, Fig.~\ref{taxi_temporal_mode_2_pickup} shows both the temporal mode 2 and two trip time series curves of the highlighted zones (i.e., zones 161 and 237) in the spatial mode 2. The temporal mode 2 is informative before the COVID-19 pandemic and shows consistent patterns with the daily trips of zones 161 and 237.\footnote{The unique zone identifiers are generated by the NYC Taxi and Limousine Commission (TLC), and the taxi zone map is available at \url{https://www1.nyc.gov/assets/tlc/images/content/pages/about/taxi_zone_map_manhattan.jpg}.} Fig.~\ref{taxi_temporal_mode_4_pickup} shows both the temporal mode 4 and three trip time series curves of the highlighted zones (i.e., zones 107, 231, and 234) in the spatial mode 4. The temporal mode 4 reveals the patterns of trips before and during the COVID-19 pandemic, but it shows a great change in March 2020. Fig.~\ref{taxi_temporal_mode_5_pickup} shows both the temporal mode 5 and two trip time series curves of the highlighted zones (i.e., zones 170 and 186). The temporal mode 5 is consistent with the daily trips of zones 170 and 186, and it also reveals the change of trips in March 2020.



In terms of the dropoff trips, Figure~\ref{taxi_temporal_mode_2_dropoff} shows both the temporal mode 2 and three trip time series curves of the highlighted zones (i.e., zones 161, 236, and 237). The temporal mode 2 reveals the (weekly) seasonal patterns of trips before the crisis of COVID-19. It is consistent with the daily trips of these zones, showing strong quasi-seasonality before the COVID-19 pandemic. It is clear to see the difference of trip patterns before and during the COVID-19 pandemic. Figure~\ref{taxi_temporal_mode_4_dropoff} shows both the temporal mode 4 and three trip time series curves of the highlighted zones (i.e., zones 79, 234, and 249). The temporal mode 4 demonstrates the trip patterns that are consistent with daily trips of these zones. Figure~\ref{taxi_temporal_mode_5_dropoff} shows both the temporal mode 5 and the trip time series curve of highlighted zone 246. The temporal mode 5 is consistent with the daily trips of zone 246 before the COVID-19 pandemic, and the pattern changes since March 2020. Remarkable traffic/trip reduction has been reported due to the travel restriction during the COVID-19 pandemic.



\section{Conclusion}\label{conclusion}

Spatiotemporal data are increasingly accessible due to the remarkable development of sensing technologies and advanced information systems. These data provide unprecedented opportunities for discovering complex dynamic mechanisms and data patterns from nonlinear systems, and it is essential to understand them through data-driven approaches. Introducing reduced-rank VAR models such as DMD into spatiotemporal data has been demonstrated to be efficient for discovering spatiotemporal patterns \cite{tu2013dynamic,kutz2016dynamic}; however, these models are sensitive to practical data noises \cite{wu2021challenges} and incapable of characterizing the time-varying system behaviors of data. Therefore, it still demands data-driven frameworks that are well-suited to spatiotemporal data.

To this end, this work presents a time-varying reduced-rank VAR model for discovering interpretable modes from time series, providing insights into modeling real-world time-varying spatiotemporal systems. The model is built on the time-varying VAR and allows one to compress the over-parameterized coefficients by low-rank tensor factorization. Experiments demonstrated that the model can reveal meaningful spatial and temporal patterns underlying the time series through interpretable modes. To evaluate the performance, we test our model on several real-world spatiotemporal datasets, including fluid dynamics, SST, USA surface temperature, and NYC taxi trips. Our model can discover the underlying spatial patterns through the spatial modes and identify the time-varying system behaviors through the temporal modes. Last but not least, it would be of interest to develop more general time-varying autoregression models for spatiotemporal data in the presence of high-dimensional time series. 


%

\appendices

\section{DMD on Fluid Dynamics}\label{appendix_dmd}

\begin{figure}[ht!]
    \centering
    \includegraphics[width=0.35\textwidth]{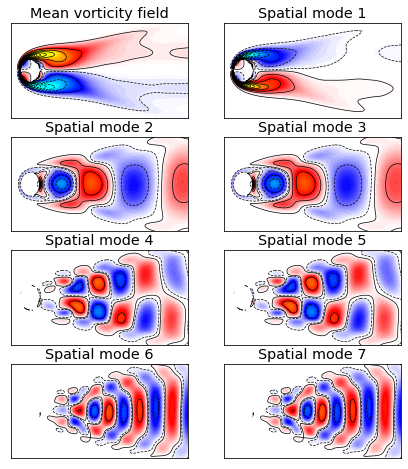}
    \caption{Mean vorticity field and spatial modes of the fluid flow achieved by DMD with the rank $R=7$. Note that the colorbars of all modes are on the same scale.}
    \label{fluid_mode_dmd}
\end{figure}

\begin{figure}[ht!]
    \centering
    \includegraphics[width=0.35\textwidth]{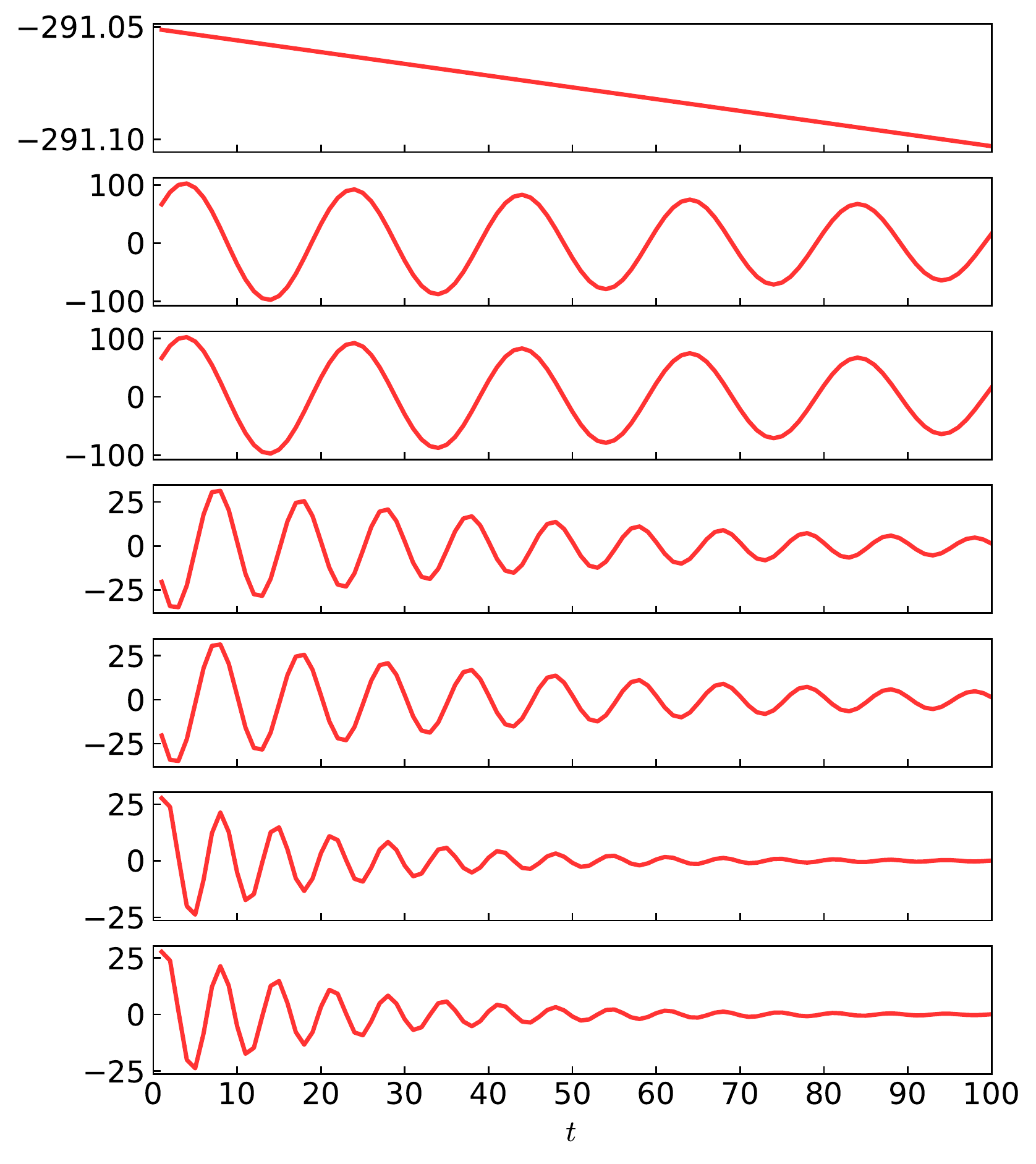}
    \caption{Temporal modes of the fluid flow achieved by DMD. Seven panels correspond to the rank $R=7$.}
    \label{fluid_temporal_mode_dmd}
\end{figure}

To highlight the advantages of the proposed model for discovering interpretable modes from spatiotemporal data, we consider DMD \cite{tu2013dynamic, kutz2016dynamic} as an important baseline for comparison. Fig.~\ref{fluid_mode_dmd} and \ref{fluid_temporal_mode_dmd} show the spatial modes and the temporal modes achieved by the DMD model, respectively. In contrast to the temporal modes achieved by our model (see Fig.~\ref{fluid_temporal_mode}), the temporal modes achieved by DMD as shown in Fig.~\ref{fluid_temporal_mode_dmd} is hard to reveal the time-varying system behaviors, failing for the changing dynamics in this case. Comparing to DMD, multiresolution DMD proposed by \cite{kutz2016multiresolution} still requires a well designed scheme, and it requires fixed frequency in each window. Going back to our model, it is fully time-varying, making it more flexible than the DMD models.

\ifCLASSOPTIONcompsoc
  \section*{Acknowledgments}
\else
  \section*{Acknowledgment}
\fi

X. Chen and C. Zhang would like to thank the Institute for Data Valorisation (IVADO) and the Interuniversity Research Centre on Enterprise Networks, Logistics and Transportation (CIRRELT) for providing the PhD Excellence Scholarship to support this study. We also thank Dr. Wenshuo Wang for providing insightful suggestions for improving this work.

\ifCLASSOPTIONcaptionsoff
  \newpage
\fi



%

\bibliographystyle{IEEEtran}
\bibliography{references}

%

\begin{IEEEbiography}[{\includegraphics[width=1in,height=1.25in,clip,keepaspectratio]{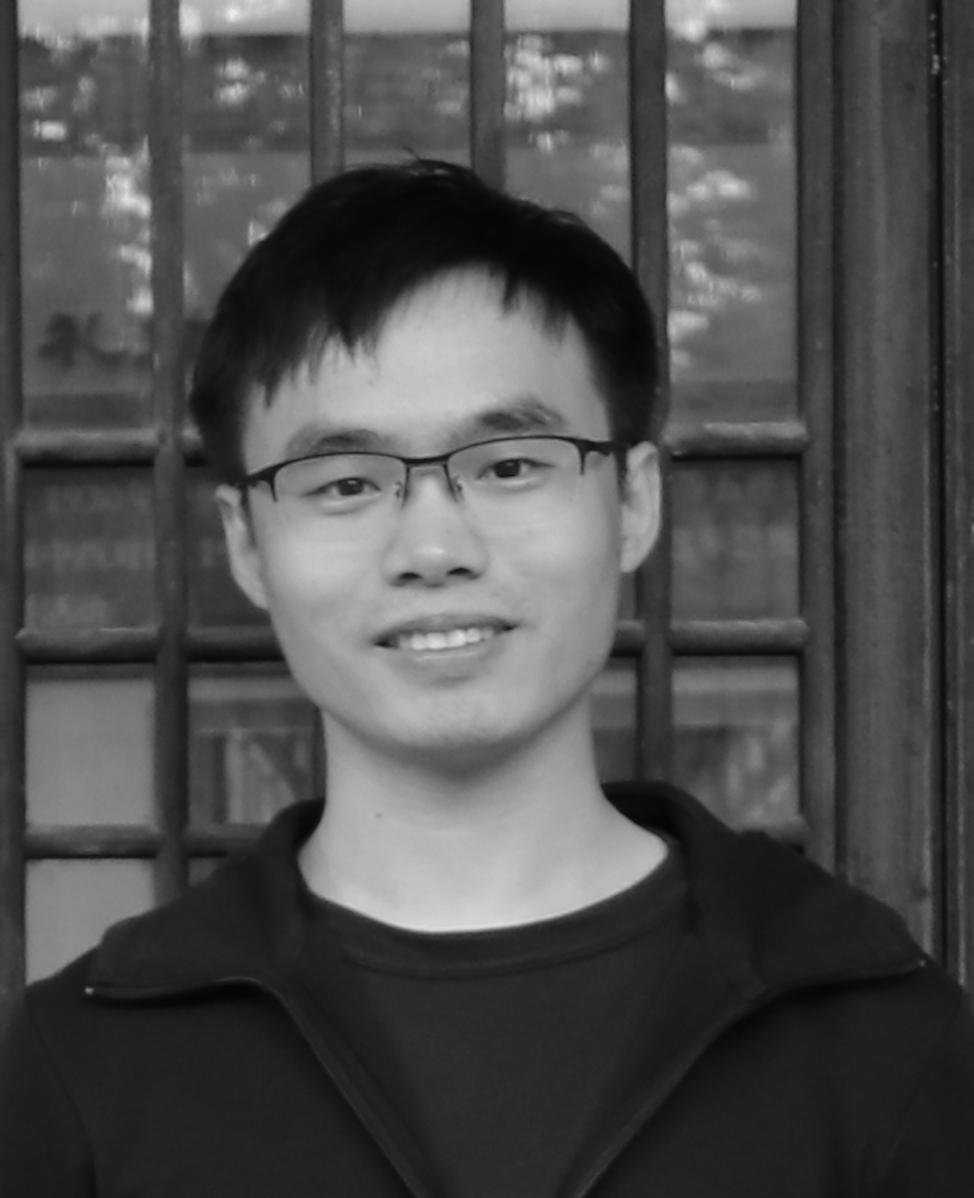}}]{Xinyu Chen}(Student Member, IEEE) received the B.S. degree in Traffic Engineering from Guangzhou University, Guangzhou, China, in 2016, and M.S. degree in Transportation Information Engineering \& Control from Sun Yat-Sen University, Guangzhou, China, in 2019. He is currently a Ph.D. student with the Civil, Geological and Mining Engineering Department, Polytechnique Montreal, Montreal, QC, Canada. His current research centers on machine learning, spatiotemporal data modeling, and intelligent transportation systems.
\end{IEEEbiography}


\begin{IEEEbiography}[{\includegraphics[width=1in,height=1.25in,clip,keepaspectratio]{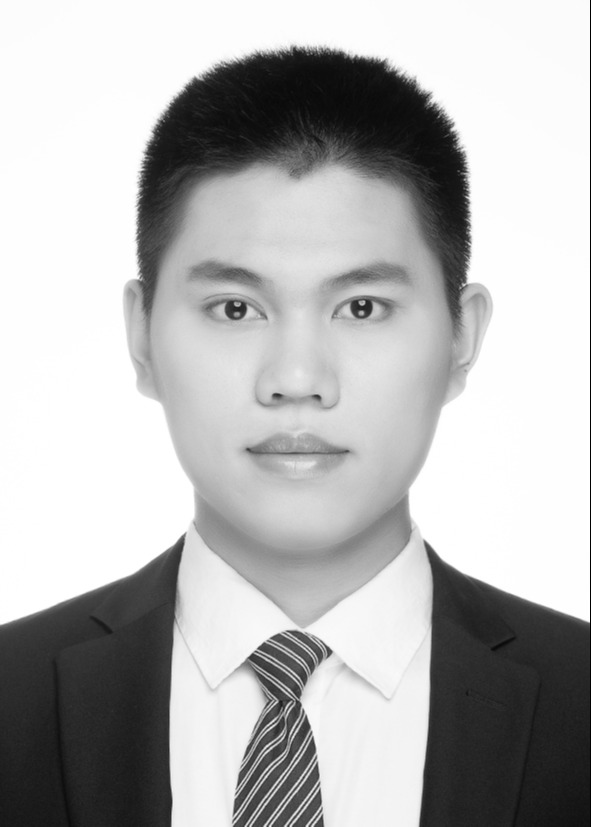}}]{Chengyuan Zhang}(Student Member, IEEE) is a Ph.D. student with the Department of Civil Engineering, McGill University, Montreal, QC, Canada. He received his B.S. degree in Vehicle Engineering from Chongqing University in 2019. From 2019 to 2020, he was as a visiting researcher with the Department of Mechanical Engineering, UC Berkeley. His research interests are Bayesian learning, macro/micro driving behavior analysis, and multi-agent interaction modeling in intelligent transportation systems.
\end{IEEEbiography}

\begin{IEEEbiography}[{\includegraphics[width=1in,height=1.25in,clip,keepaspectratio]{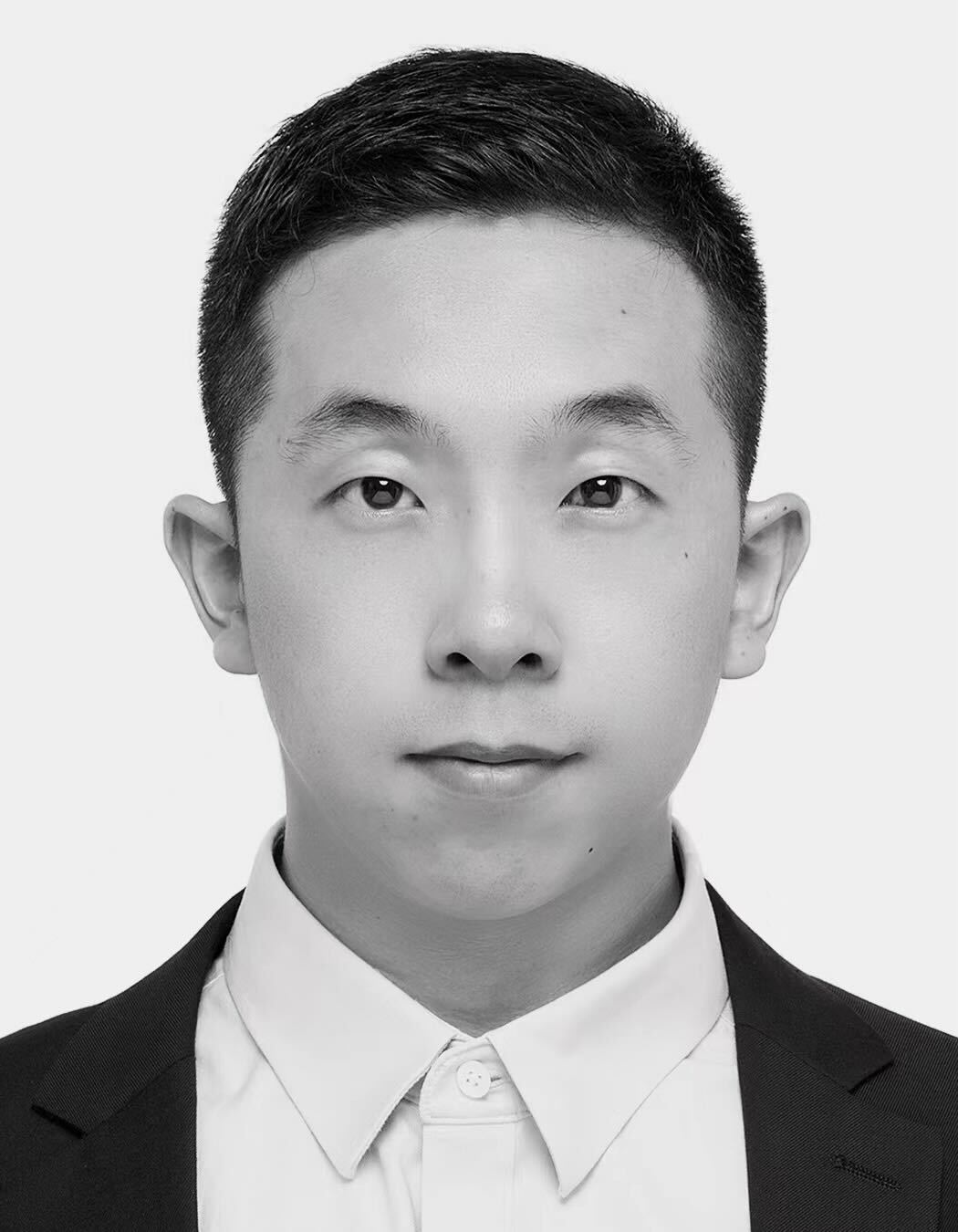}}]{Xiaoxu Chen} is a Ph.D. student with the Department of Civil Engineering, McGill University, Montreal, QC, Canada. He received the B.S. degree in transportation engineering from Harbin Institute of Technology in 2017 and the M.S. degree in transportation engineering from Tongji University, Shanghai, China, in 2020. His research currently focuses on Bayesian statistics, spatiotemporal transportation modeling.
\end{IEEEbiography}

\begin{IEEEbiography}[{\includegraphics[width=1in,height=1.25in,clip,keepaspectratio]{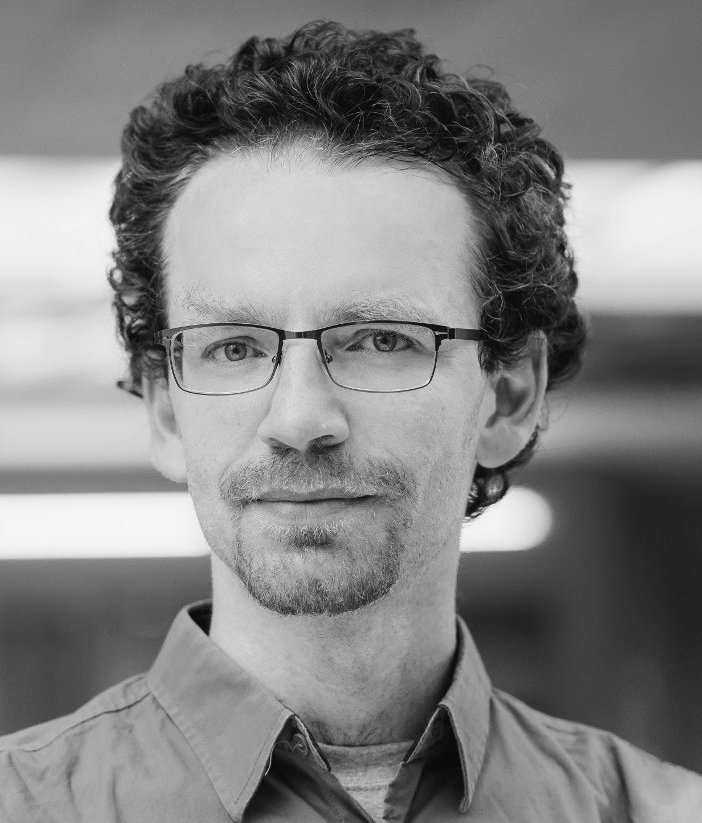}}]{Nicolas Saunier} received an engineering degree and a Doctorate (Ph.D.) in computer science from Telecom ParisTech, Paris, France, respectively in 2001 and 2005. He is currently a Full Professor with the Civil, Geological and Mining Engineering Department at Polytechnique Montreal, Montreal, QC, Canada. His research interests include intelligent transportation, road safety, and data science for transportation.
\end{IEEEbiography}

\begin{IEEEbiography}[{\includegraphics[width=1in,height=1.25in,clip,keepaspectratio]{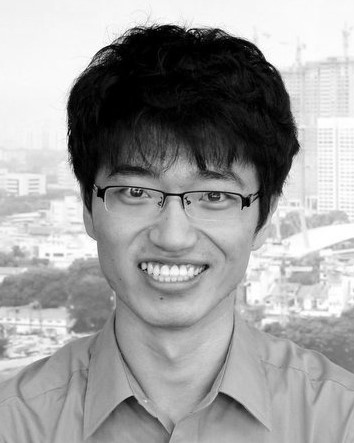}}]{Lijun Sun}(Senior Member, IEEE) received the B.S. degree in civil engineering from Tsinghua University, Beijing, China, in 2011, and the Ph.D. degree in civil engineering (transportation) from the National the University of Singapore in 2015. He is currently an Assistant Professor in the Department of Civil Engineering, McGill University, Montreal, QC, Canada. His research centers on intelligent transportation systems, machine learning, spatiotemporal modeling, travel behavior, and agent-based simulation.
\end{IEEEbiography}


\vfill


\end{document}